\documentclass[journal]{IEEEtran}
\ifCLASSINFOpdf

\else

\fi

\usepackage{mathrsfs}
\usepackage{amsmath}
\usepackage{amsfonts}
\usepackage{amsthm}
\usepackage{amssymb}
\usepackage{graphicx}
\usepackage{subfigure}
\usepackage{indentfirst}
\usepackage{array}
\usepackage{epstopdf}
\usepackage{cite}
\usepackage{mathrsfs}
\usepackage{enumerate}
\usepackage{bm}
\usepackage[linesnumbered,ruled,vlined]{algorithm2e}
\usepackage{setspace}
\usepackage{multirow}
\usepackage{verbatim}
\usepackage{stfloats}
\usepackage{color}
\usepackage[table]{xcolor}
\usepackage{makecell}
\usepackage{cancel}

\SetKwComment{Comment}{//}{}
\SetKwBlock{Begin}{Function}{end}

\begin{document}
\title{Wireless Deep Video Semantic Transmission}

\author{Sixian~Wang,~\IEEEmembership{Graduate Student Member,~IEEE},
Jincheng~Dai,~\IEEEmembership{Member,~IEEE},
Zijian~Liang,
Kai~Niu,~\IEEEmembership{Member,~IEEE},
Zhongwei~Si,~\IEEEmembership{Member,~IEEE},
Chao~Dong,~\IEEEmembership{Member,~IEEE},
Xiaoqi~Qin,~\IEEEmembership{Member,~IEEE},
and Ping~Zhang,~\IEEEmembership{Fellow,~IEEE}

\thanks{This work was supported in part by the National Natural Science Foundation of China under Grant 92067202, Grant 62001049, Grant 62071058, Grant 61971062, in part by the Beijing Natural Science Foundation under Grant 4222012, and in part by the Major Key Project of PCL under Grant PCL2021A15. (\emph{Corresponding author: Jincheng Dai})}

\thanks{S. Wang, J. Dai, Z. Liang, K. Niu, Z. Si, C. Dong are with the Key Laboratory of Universal Wireless Communications, Ministry of Education, Beijing University of Posts and Telecommunications, Beijing 100876, China (corresponding author e-mail: daijincheng@bupt.edu.cn).}

\thanks{K. Niu is also with Peng Cheng Laboratory, Shenzhen, China.}

\thanks{X. Qin and P. Zhang are with the State Key Laboratory of Networking and Switching Technology, Beijing University of Posts and Telecommunications, Beijing 100876, China.}

\vspace{0em}
}

\maketitle

\begin{abstract}
In this paper, we design a new class of high-efficiency deep joint source-channel coding methods to achieve end-to-end video transmission over wireless channels. The proposed methods exploit nonlinear transform and conditional coding architecture to adaptively extract semantic features across video frames, and transmit semantic feature domain representations over wireless channels via deep joint source-channel coding. Our framework is collected under the name deep video semantic transmission (DVST). In particular, benefiting from the strong temporal prior provided by the feature domain context, the learned nonlinear transform function becomes temporally adaptive, resulting in a richer and more accurate entropy model guiding the transmission of current frame. Accordingly, a novel rate adaptive transmission mechanism is developed to customize deep joint source-channel coding for video sources. It learns to allocate the limited channel bandwidth within and among video frames to maximize the overall transmission performance. The whole DVST design is formulated as an optimization problem whose goal is to minimize the end-to-end transmission rate-distortion performance under perceptual quality metrics or machine vision task performance metrics. Across standard video source test sequences and various communication scenarios, experiments show that our DVST can generally surpass traditional wireless video coded transmission schemes. The proposed DVST framework can well support future semantic communications due to its video content-aware and machine vision task integration abilities.
\end{abstract}


\IEEEpeerreviewmaketitle

\section{Introduction}\label{section_introduction}

\IEEEPARstart{T}{he} task of video transmission in today's wireless networks is largely separated into two steps: source coding and channel coding \cite{Shannon1948A}. Source coding compresses the source video as sequences of bits, and channel coding represents sequences of bits as transmitted signals against impacts of imperfect wireless channels such as noise, fading, and interferences. This separation-based approach has been employed for a large variety of applications, as the binary representations of various source data can be seamlessly transmitted over arbitrary wireless channels by changing the underlying channel code. This paradigm has benefited a lot due to the independent optimization of each component.

However, the limits of the separation-based design begin to emerge with more demands on low-latency wireless video delivery applications such as virtual reality (VR). On the one hand, current wireless video transmission systems suffer from time-varying channel conditions, in which case the mismatch between communication rate and channel capacity leads to obvious \emph{cliff-effect}, i.e., the performance breaks down when the channel capacity goes below communication rate. On the other hand, the widely-used entropy coding, which converts the source representation into sequences of bits, is quite sensitive to the variational estimate of the marginal distribution of the source latent representation. Small perturbations on this marginal can lead to catastrophic error propagation in entropy decoding \cite{rissanen1981universal}. In practice, the small perturbation can often be caused by floating point round-off error \cite{balle2018integer}. Unfortunately, this round-off operation depends heavily on hardware or software platforms, and in various data compression applications, the transceiver may well employ different platforms. This non-determinism issue in transmitter vs. receiver may finally lead to severe performance degradation.

To address this problem, it is very time to bridge source coding and channel coding together to boosting the end-to-end communication system capabilities. By this means, the channel transmission process can be aware of the source semantic features \cite{zhang2021toward,xie2021deep,qin2021semantic,seo2021semantics,dai2021semantic,dai2021nonlinear,niu2022towards}. The paradigm aiming at the integrated design of source and channel processing was named \emph{joint source-channel coding (JSCC)} \cite{verduJSCC}, a classical topic in the information theory and coding theory. However, conventional JSCC schemes \cite{verduJSCC,guyader2001joint,ramzan2007joint,chen2018joint} are based on explicit probabilistic models and handcrafted designs, whose optimization complexity is intractable for complex sources. In addition, they ignored the semantic aspects of source messages. As one modern version, recent deep learning methods for realizing JSCC have stimulated significant interest in both artificial intelligence (AI) and wireless communication communities \cite{JSCCtext,choi2019neural,DJSCC,DJSCCL,jankowski2020wireless}. By using artificial neural networks (ANNs), source data can be directly encoded as continuous-valued symbols to be transmitted over wireless channels. Deep JSCC can overcome the catastrophic degradation problem by using analog transmission without entropy coding. Current deep JSCC methods have shown end-to-end image transmission performance surpassing classical separation-based JPEG/JPEG2000/BPG source compression combined with ideal channel capacity-achieving code family, especially for sources of small dimensions, e.g., small CIFAR10 image data set \cite{krizhevsky2009learning}.

However, one can observe that, in general, as the source dimension increases, e.g., large-scale images, the performance of deep JSCC degrades rapidly, which is even inferior to the classical separation-based coding schemes as demonstrated in \cite{dai2021nonlinear}. Moreover, existing deep JSCC schemes cannot provide comparable coding gain as that of classical separated coding schemes, i.e., the slope of the performance curve slows down quickly with the increase of coding rate or the channel signal-to-noise ratio (SNR). This poor coding gain stems from the naive design of codec networks. Current deep JSCC works simply employ one highly-integrated ANN as the encoder function to achieve dimension reduction with respect to the raw source data. By adding the wireless channel as one non-trainable layer, the learned codec ANNs can also combat the impacts of imperfect wireless channels. Nevertheless, this light auto-encoder structured deep JSCC cannot provide sufficient model expression capability for large-scale source data, resulting in a prematurely saturated coding gain of deep JSCC. Compared to the image source, the video source further induces the time dimension. Above saturation phenomenon on the coding gain is more likely to appear on video sources that need higher dimensional representation. Thus, a naive application of deep JSCC for wireless video transmission cannot provide satisfactory performance.

Inspired by the emerging data compression methods adopted in computer vision (CV) communities, a high-dimensional source will first be converted as latent representations defined by variational latent-variable models. This procedure is referred to as \emph{nonlinear transform} \cite{balle2017,balle2018efficient,balle2018,minnen2018,balle2020nonlinear}. The richness of latent representations preserves almost all the source semantic features that can be used for either recovering source data or directly driving the downstream intelligent tasks. By training appropriately, many such nonlinear transform models successfully represent the source data quite compactly and may be called compression in a sense. For practical data compression tasks, the latent representation needs to be further compressed as binary sequences through entropy coding. However, because the employed ANNs in nonlinear transform are typically based on floating point math, and the transmission is over time-varying wireless channels, a direct combination of nonlinear transform with traditional source coding (entropy coding such as arithmetic coding \cite{rissanen1981universal}) and channel coding (such as low-density parity-check (LDPC) coding \cite{richardson2018design}) will be also vulnerable to catastrophic failures left by entropy decoding.

In this paper, to attain high-efficiency and robust end-to-end video transmission, we leverage the advantages of nonlinear transform and deep JSCC together to formulate a new powerful framework, named deep video semantic transmission (DVST). It is specifically targeted at video transmission over imperfect wireless channels and preventing catastrophic failures caused by sensitive entropy coding. By integrating the emerging conditional coding paradigm \cite{DCVC} with nonlinear transform and deep JSCC, the proposed DVST framework works on the principle: considering the strong temporal correlations among video frames, DVST encodes the current frame in an efficient manner to generate channel-input symbols through \emph{contextual nonlinear transform} and \emph{contextual deep JSCC}. The contextual semantic information is used as part of the input of both nonlinear transform and deep JSCC codec. Benefiting from the temporal prior provided by the semantic feature domain context, the learned nonlinear transform function becomes temporally adaptive, resulting in a richer and more accurate entropy model to indicate how to allocate channel bandwidth resources to transmit the current frame. Moreover, we leverage the context to carry rich information as prior to deep JSCC codec, which helps to reconstruct the semantic feature map for higher video quality or downstream task performance. The whole DVST design is formulated as an optimization problem whose goal is to minimize the end-to-end transmission rate-distortion (RD) performance under perceptual quality metrics or machine vision task performance metrics.

Specifically, the contributions of this paper can be summarized as follows.

\begin{enumerate}[(1)]
\item \emph{DVST Framework:} We propose a new end-to-end learnable framework for wireless video transmission, i.e., DVST, which integrates the advantages of nonlinear transform and deep JSCC. To the best of authors' knowledge, this is the first work that establishes a temporally adaptive entropy model to customize deep JSCC for video. The proposed DVST framework exploits both nonlinear transform and conditional coding architecture for video semantic feature extraction, which contributes to higher efficiency and more robust wireless video transmission than traditional video coded transmission schemes.

\item \emph{Context-Driven Semantic Feature Modeling:} We exploit a simple yet efficient method using temporal context to enhance the entropy model in nonlinear transform as well as the deep JSCC codec. We design ANN architectures to realize each module of DVST, in which the definition, usage, and learning manner of contextual semantic features as conditions are all clearly given.

\item \emph{Rate-Adaptive Semantic Feature Transmission:} In light of the temporally adaptive entropy model on semantic features, we develop a method to improve the coding gain of video deep JSCC. In particular, we introduce a variable-length transmission mechanism for each embedding vector in the latent representation. The resulting DVST model learns to allocate the limited channel bandwidth within and among video frames to maximize the overall performance.

\item \emph{Performance Validation:} We verify the performance of our DVST system across standard video source sequences. We show that for wireless video transmission, our DVST can achieve much better coding gain and RD performance on various established metrics such as PSNR and MS-SSIM. Equivalently, achieving identical end-to-end wireless transmission performance, the proposed DVST method can save up to 50\% channel bandwidth cost, compared to classical H.264/H.265 combined with LDPC and digital modulation schemes. For task-oriented machine-type semantic communications, experimental results verify the effectiveness of DVST, which can better support machine vision tasks, meanwhile holds higher perceptual fidelity for human vision.
\end{enumerate}

The remainder of this paper is organized as follows. In the next section \ref{section_framework}, we first review the system model of wireless video transmission, and propose the DVST framework. Then, in section \ref{section_architecture}, we propose ANN architectures for realizing DVST, as well as key methods to guide the optimization of the DVST model. Section \ref{section_results} provides a direct comparison of a number of methods to quantify the performance gain of the proposed method. Finally, section \ref{section_conclusion} concludes this paper.

\begin{figure*}[t]
\setlength{\abovecaptionskip}{0.cm}
\setlength{\belowcaptionskip}{-0.cm}
\centering{\includegraphics[scale=0.4]{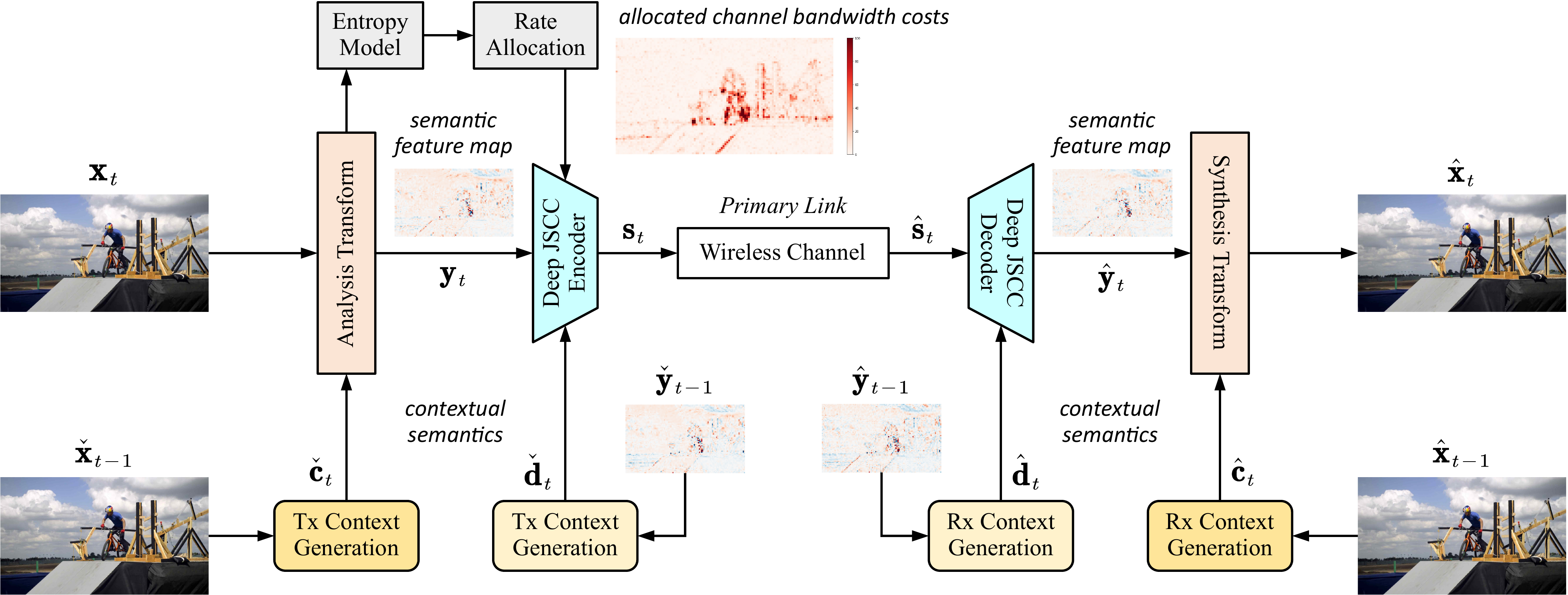}}
\caption{The framework of DVST. The visualization of the latents with the highest entropy.}\label{Fig1}
\vspace{0em}
\end{figure*}

\emph{Notational Conventions:} Throughout this paper, lowercase letters (e.g., $x$) denote scalars, bold lowercase letters (e.g., $\mathbf{x}$) denote vectors. In some cases, $x_i$ denotes the elements of $\mathbf{x}$, which may also represent a subvector of $\mathbf{x}$ as described in the context. Bold uppercase letters (e.g., $\mathbf{X}$) denote matrices, and $\mathbf{I}_m$ denotes an $m$-dimensional identity matrix. $\ln (\cdot)$ denotes the natural logarithm, and $\log (\cdot)$ denotes the logarithm to base $2$. $p_x$ denotes a probability density function (pdf) with respect to the continuous-valued random variable $x$, and $P_{\bar x}$ denotes a probability mass function (pmf) with respect to the discrete-valued random variable $\bar x$. In addition, $\mathbb{E} [\cdot]$ denotes the statistical expectation operation, and $\mathbb{R}$ denotes the real number set. Finally, $\mathcal{N}(x|\mu, \sigma^2) \triangleq (2\pi \sigma^2)^{-1/2} \exp(-(x - \mu)^2/(2\sigma^2))$ denotes a Gaussian function, ${\mathcal{L}}(x|\mu, \sigma) \triangleq (2\sigma)^{-1} \exp(-|x - \mu|/\sigma)$ denotes a Laplace function, and $\mathcal{U}(a-u,a+u)$ stands for a uniform distribution centered on $a$ with the range from $a-u$ to $a+u$.

\section{The Proposed Method}\label{section_framework}

In this section, we first present the system model of wireless video transmission. Then, we describe the whole framework of DVST. After that, we introduce the contextual entropy model for rate-adaptively transmit the latent representations, followed by the learning methods of the context. Finally, we derive the optimization goal of DVST system.

\subsection{System Model}

Consider a wireless video transmission problem. Given a video sequence $\mathcal{X} = \left\lbrace \mathbf{x}_1, \mathbf{x}_2, \dots , \mathbf{x}_T \right\rbrace$, where the frame at time step $t$ is modeled as a vector of pixel intensities $\mathbf{x}_t \in \mathbb{R}^{m}$. The transmitter encodes the video frame sequence $\mathcal{X}$ as a sequence of variable-length continuous-valued channel input symbols $\mathcal{S} = \left\lbrace \mathbf {s}_1, \mathbf {s}_2, \dots, \mathbf{s}_T \right\rbrace$, where $\mathbf{s}_t \in \mathbb{R}^{k_t}$ denote the $k_t$-dimensional channel input vector at time step $t$. We usually have $k_t < m$, and $R = \frac{1}{T} \sum_{t=1}^{T} \frac{k_{t}}{m}$ is defined as the \emph{channel bandwidth ratio (CBR)} \cite{DJSCCF} denoting the average coding rate of $\mathcal{X}$. Then, the sequence $\{\mathbf {s}_t \}$ is successively sent over the wireless channel. This channel introduces random corruptions denoted as a transfer function $W(\cdot;\bm{\nu})$, where $\bm{\nu}$ denotes the channel parameters. The received sequence is ${\mathbf{\hat s}}_t = W(\mathbf{s}_t;\bm{\nu})$ with the transition probability $p_{{\mathbf{\hat s}}_t|\mathbf {s}_t}({\mathbf{\hat s}}_t|\mathbf{s}_t)$. In this paper, we mainly consider the widely used additive white Gaussian noise (AWGN) channel such that the transfer function is ${\mathbf{\hat s}}_t = W( \mathbf{s}_t ; \sigma^2 ) = \mathbf{s}_t + \mathbf{n}_t$ where each component of the noise vector $\mathbf{n}_t$ is independently sampled from a time-invariant multidimensional Gaussian distribution, i.e., $\mathbf{n}_t \sim \mathcal{N}(0, {\sigma^2}{\mathbf{I}}_{k_t})$, where ${\sigma^2}$ is the average noise power. Other channel models can also be similarly incorporated by changing the channel transition function. The receiver comprises a series of inverse operation which aims to recover $\mathbf{\hat x}_t$ from the corrupted signal $\mathbf{\hat s}_t$ or executes the downstream intelligent task.

We consider video transmission over the noisy wireless channel in a low-latency manner, i.e., the video sequence is transmitted to and reconstructed in the receiver frame-by-frame. We encapsulate $N$ consecutive frames as one group-of-pictures (GOP). A typical video coding algorithm first divides $\mathcal{X}$ into a stack of GOPs. Each GOP begins with an intra-coded picture (I-frame or keyframe) as a reference, followed by $N-1$ predictive coded frames (P-frames), which contain the motion compensated difference information for bitrate saving. In this paper, we exploit the classical GOP structure for end-to-end transmission. Since the transmission of I-frame is equivalent to that of image, which has been well studied in \cite{DJSCC,DJSCCF,DJSCCL,jankowski2020wireless}, we concentrate on the transmission of P-frame.

\subsection{The Framework of DVST}

We propose the DVST as a new learnable model for end-to-end wireless video transmission, which integrates the advantages of nonlinear transform and deep JSCC. Our DVST framework is illustrated in Fig. \ref{Fig1}. To encode the current frame ${\mathbf x}_t$ efficiently, the transmitter adopts \emph{contextual analysis transform} and \emph{contextual deep JSCC encoder} as two critical modules. The analysis transform converts the source frame in pixel domain to the latent representation in semantic feature domain. Guided by the variational entropy modeling on the latent representation, a rate control module is added to achieve variable-length encoding in deep JSCC. For video source, there exists temporal correlation. Thus, the above two modules also employ the semantic feature domain context and the deep JSCC codeword domain context as the temporal prior. That makes nonlinear transform and deep JSCC modules temporally adaptive, resulting in a higher efficient video transmission framework.

\begin{figure}[t]
\setlength{\abovecaptionskip}{0.cm}
\setlength{\belowcaptionskip}{-0.cm}
\centering{\includegraphics[scale=0.4]{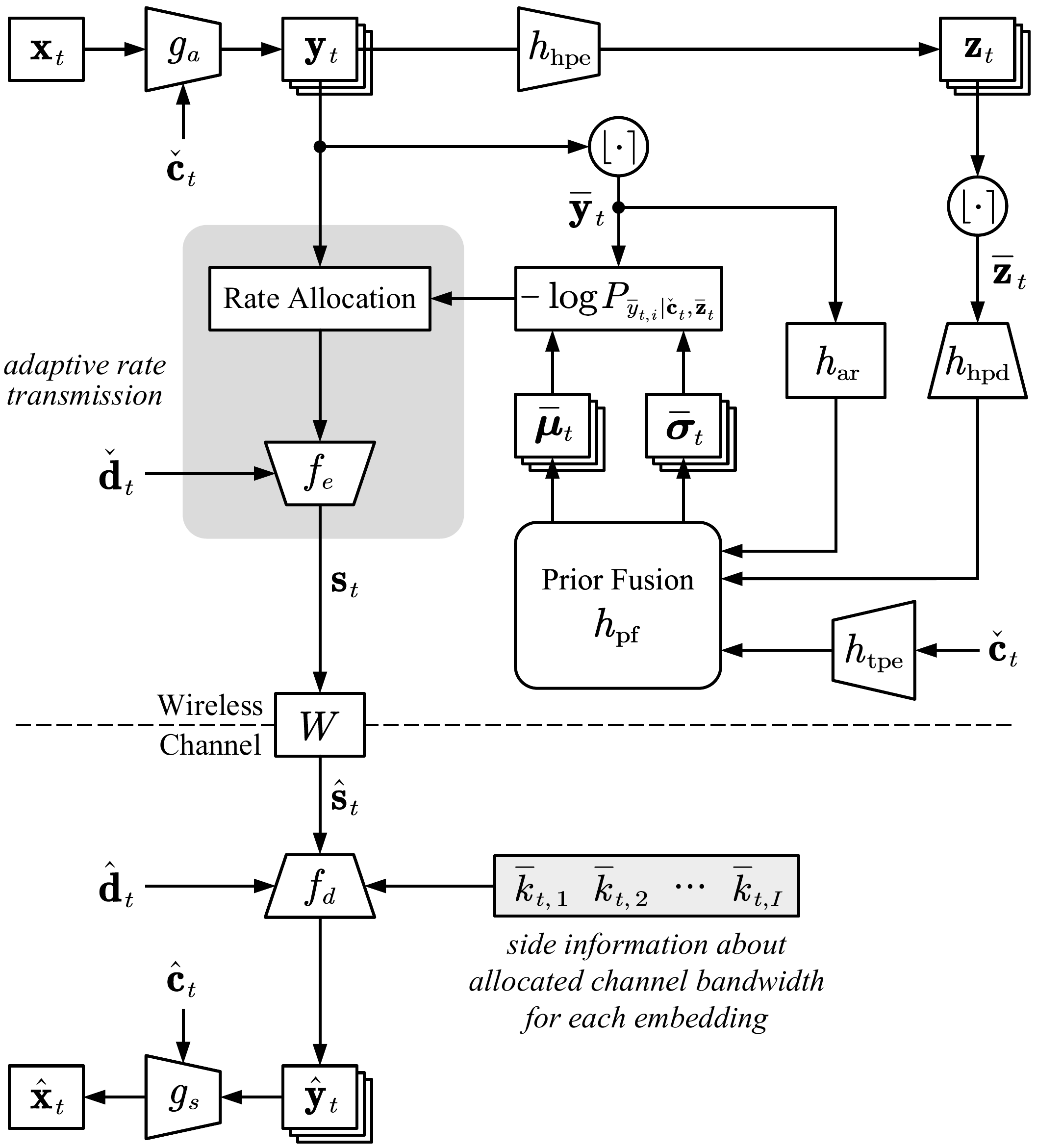}}
\caption{Illustration of a DVST codec architecture and the entropy model of the primary link.}\label{Fig2}
\vspace{-0.5em}
\end{figure}

As illustrated in Fig. \ref{Fig1} and Fig. \ref{Fig2}, $g_a$ and $g_s$ functions in nonlinear transform of the current frame are conditioned on the contextual semantic features ${\mathbf{\check c}}_t$ and ${\mathbf{\hat c}}_t$, respectively. $f_e$ and $f_d$ functions in deep JSCC are conditioned on the contextual codewords ${\mathbf{\check d}}_t$ and ${\mathbf{\hat d}}_t$, respectively. The \emph{primary link} of DVST system is formulated as
\begin{equation}\label{eq_video_trans_process}
{{\mathbf{x}}_t} \xrightarrow{{{g_a}( \cdot |{\mathbf{\check c}}_t)}} {{\mathbf{y}}_t}
\xrightarrow{{{f_e}( \cdot |{\mathbf{\check d}}_t)}} {{\mathbf{s}}_t}
\xrightarrow{{{W}( \cdot |\bm{\nu}})} {{\mathbf{\hat s}}_t}
\xrightarrow{{{f_d}( \cdot |{\mathbf{\hat d}}_t)}} {{\mathbf{\hat y}}_t}
\xrightarrow{{{g_s}( \cdot |{\mathbf{\hat c}}_t)}} {{\mathbf{\hat x}}_t}.
\end{equation}
In our DVST design, we use ANN to realize each function in \eqref{eq_video_trans_process} except for the channel transfer function $W$. In terms of the context information, transmitter (Tx) contexts ${\mathbf{\check c}}_t$ and ${\mathbf{\check d}}_t$ are obtained from the reference frame ${\mathbf{\check x}}_{t-1}$ and the reference feature map ${\mathbf{\check y}}_{t-1}$, respectively. These two references are generated at the transmitter by simulating the DVST process without passing over the wireless channel, i.e.,
\begin{equation}\label{eq_video_trans_process_simulate}
{{\mathbf{s}}_{t-1}}
\xrightarrow{{{f_d}( \cdot |{\mathbf{\check d}}_{t-1})}} {{\mathbf{\check y}}_{t-1}}
\xrightarrow{{{g_s}( \cdot |{\mathbf{\check c}}_{t-1})}} {{\mathbf{\check x}}_{t-1}},
\end{equation}
where the codeword ${{\mathbf{s}}_{t-1}}$ is obtained from \eqref{eq_video_trans_process} by substituting the time step $t-1$. Receiver (Rx) contexts ${\mathbf{\hat c}}_t$ and ${\mathbf{\hat d}}_t$ are obtained from the reference synthesized frame ${\mathbf{\hat x}}_{t-1}$ and the reference decoded feature map ${\mathbf{\hat y}}_{t-1}$, respectively. These two references are directly obtained by taking out records from the receiver buffer at the time step $t-1$. Details about how to use context as conditions to formulate ANN-based functions will be introduced in the next section.

Specifically, in the transmitter, for the current frame ${\mathbf x}_t$ at time step $t$, $g_a(\cdot)$ extracts the source semantic features as a lower-dimensional latent representation ${\mathbf{ y}_t}$, $f_e(\cdot)$ operates on this latent space. Consider the inter frame correlation in video sources, the analysis transform $g_a(\cdot)$ is formulated as
\begin{equation}\label{eq_context_ga}
{\mathbf{y}_t} = g_a({\mathbf x}_t | {\mathbf{\check c}}_t) \text{~with~} {\mathbf{\check c}}_t = \varphi_a( \mathbf{\check x}_{t-1}).
\end{equation}
$\varphi_a(\cdot)$ denotes the function to generate context ${\mathbf{\check c}}_t$ for the analysis transform, $g_a(\cdot)$ is therefore referred to as the \emph{contextual analysis transform}. After that, the latent representation ${\mathbf{y}_t}$ is fed into the \emph{contextual deep JSCC encoder} $f_e(\cdot)$ to generate the channel-input sequence $\mathbf{s}_t$ as
\begin{equation}\label{eq_context_fe}
{\mathbf{s}_t} = f_e({\mathbf y}_t | {\mathbf{\check d}}_t) \text{~with~} {\mathbf{\check d}}_t = \gamma_{e}( \mathbf{\check{y}}_{t-1}).
\end{equation}
$\gamma_e(\cdot)$ denotes the function of generating context for deep JSCC encoder. To provide rich and more correlated information for encoding $\mathbf{x}_t$, the context ${\mathbf{\check c}}_t$ is in the semantic feature domain with higher dimensions, and the context ${\mathbf{\check d}}_t$ is in the deep JSCC codeword space.

Then, the analog codeword sequence $\mathbf{s}$ is directly sent over the wireless communication channel. As aforementioned, we consider the AWGN channel such that the received sequence is ${\mathbf{\hat s}}_t = \mathbf{s}_t + \mathbf{n}_t$ with $\mathbf{n}_t \sim \mathcal{N}(0, {\sigma^2}{\mathbf{I}}_{k_t})$. The receiver comprises a \emph{contextual deep JSCC decoder} $f_d(\cdot)$ to reconstruct the corrupted signal ${\mathbf{\hat s}}_t$ as ${\mathbf{\hat y}}_t$, i.e.,
\begin{equation}\label{eq_context_fd}
{\mathbf{\hat{y}}_t} = f_d({\mathbf{\hat s}}_t | {\mathbf{\hat d}}_t) \text{~with~} {\mathbf{\hat d}}_t = \gamma_{d}( \mathbf{\hat{y}}_{t-1}).
\end{equation}
$\gamma_d(\cdot)$ denotes the function of generating context for deep JSCC decoder. The \emph{contextual synthesis transform} function $g_s(\cdot)$ is then performed on ${\mathbf{\hat{y}}_t}$ to recover the current frame, i.e.,
\begin{equation}\label{eq_context_gs}
{\mathbf{\hat x}_t} = g_s({\mathbf{\hat y}}_t | {\mathbf{\hat c}}_t) \text{~with~} {\mathbf{\hat c}}_t = \varphi_s( \mathbf{\hat x}_{t-1}).
\end{equation}
$\varphi_s(\cdot)$ denotes the function to generate context ${\mathbf{\hat c}}_t$ for the synthesis transform.

For the contextual analysis transform $g_a$, we use a network to automatically learn the correlation between $\mathbf{x}_t$ and $\mathbf{\check c}_t$, which does not remove the redundancy through handcrafted subtraction operation like traditional residual video coding \cite{lu2019dvc}. Herein, the context $\mathbf{\check c}_t$ comes from the reference frame $\mathbf{\check x}_{t-1}$. In this way, the contextual analysis transform becomes adaptive which generates the latent representation by selectively extracting semantic features from $\mathbf{x}_t$ and $\mathbf{\check x}_{t-1}$ \cite{DCVC}. Due to the motion in video, for old contents in $\mathbf{x}_t$ that can find a good reference from $\mathbf{\check x}_{t-1}$, $g_a$ still forces to generate its patch embeddings from the residue. For new contents in $\mathbf{x}_t$ that cannot find a good reference from $\mathbf{\check x}_{t-1}$, $g_a$ tends to generate its patch embeddings from $\mathbf{x}_t$ itself. The contextual nonlinear transform inherently learns to adaptively utilize the condition for semantic extraction. In addition, the context $\mathbf{\check c}_t$ is not only used for generating the latent representation, but also utilized to construct the entropy model, which will be introduced in the subsequent subsection.

For the contextual deep JSCC encoder $f_e$, we use a network to automatically learn the correlation between $\mathbf{y}_t$ and $\mathbf{\check d}_t$. Note that the context $\mathbf{\check d}_t$ comes from the reconstructed reference feature map $\mathbf{\check y}_{t-1}$, thus, the contextual deep JSCC encoder also becomes adaptive to generate the channel-input codewords. If patch embeddings in $\mathbf{y}_t$ can find a good reference from $\mathbf{\check y}_{t-1}$, $f_e$ inclines to transmit these embeddings with smaller channel bandwidth. In contrast, for patch embeddings in $\mathbf{x}_t$ that cannot find a good reference from $\mathbf{\check y}_{t-1}$, $f_e$ tends to allocate more channel bandwidth to transmit these embeddings. In this way, the contextual deep JSCC codec learns to adaptively utilize the condition for high-efficiency transmission. Moreover, the context $\mathbf{\check d}_t$ is not only used for generating the channel-input codeword, but also utilized to learn a rate-allocation function that controls the scaling rule from entropy value to channel bandwidth cost. Details will be introduced in the subsequent subsection.

\subsection{Entropy Model for Rate-Adaptive Transmission}

\begin{figure*}[t]
\setlength{\abovecaptionskip}{0.cm}
\setlength{\belowcaptionskip}{-0.cm}
\centering{\includegraphics[scale=0.4]{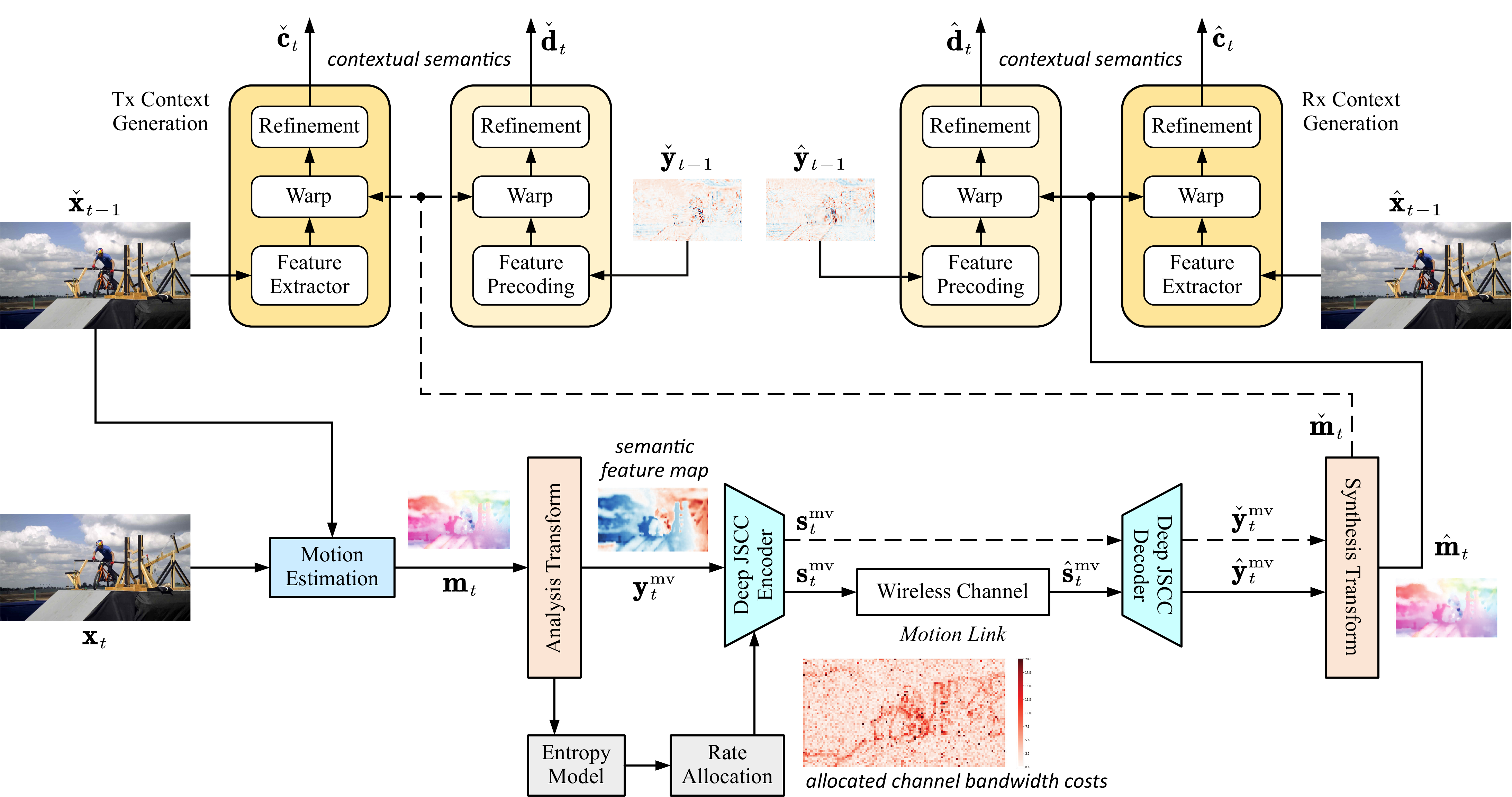}}
\caption{The framework of context learning. Dashed lines denote MV data flows simulated at the transmitter, which are used to generate the reference MV without passing over the wireless channel. Solid lines denote data flows at both transmitter and receiver, which will pass over the realistic wireless channel.}\label{Fig3}
\vspace{0em}
\end{figure*}

In order to improve the coding gain of DVST, a variable-length transmission mechanism should be developed for each embedding $y_{t,i}$ of the semantic feature map $\mathbf{y}_{t}$. To this end, we estimate the entropy distribution on $\mathbf{y}_t$, and the channel bandwidth cost $k_{t,i}$ for transmitting $y_{t,i}$ can be accordingly determined. Therefore, our target is to design an entropy model which can accurately estimate the probability distribution of the latent representation $\mathbf{y}_t$.

Our entropy model is illustrated in Fig. \ref{Fig2}. Following the work of \cite{DCVC}, the latent representation $\mathbf{y}_t$ is variationally modeled as the Laplace distribution, where each embedding $y_{t,i}$ is of varying distribution parameters. In this paper, the hyperprior entropy model learns both the hierarchical prior \cite{balle2018} and the spatial prior \cite{minnen2018}. In addition, our entropy model fuses the temporal prior provided by the context $\mathbf{\check c}_t$. Specifically, the entropy of each embedding $y_{t,i}$ is computed as
\begin{equation}\label{eq_entropy_yi}
{r_{t,i}} = -\log{P_{{\bar y}_{t,i} | \mathbf{\check c}_t, \mathbf{\bar z}_t, \mathbf{\bar y}_{t,<i} }({\bar y}_{t,i} | \mathbf{\check c}_t, \mathbf{\bar z}_t, \mathbf{\bar y}_{t,<i} )},
\end{equation}
where ${\bar y}_{t,i} = \lfloor y_{t,i} \rceil$ denotes the quantized version of $y_{t,i}$ by using the uniform scalar quantization $\lfloor \cdot \rceil$ function (rounding to integers). $\mathbf{\bar y}_{t,<i}$ denotes a tensor consisting of quantized embeddings ${\bar y}_{t,l}$ with $l<i$. The quantized hyperprior $\mathbf{\bar z}_t = \lfloor \mathbf{z}_t \rceil = \lfloor h_{\rm{hpe}}(\mathbf{y}_t) \rceil$ is obtained by stacking the hyperprior encoder network $h_{\rm{hpe}}(\cdot)$ on $\mathbf{y}_t$.

In order to use the gradient descent methods to optimize the entropy model, Ball\'{e} et al. have proposed a relaxed method for addressing the zero gradient problem caused by quantization \cite{balle2017}. A proxy ``uniformly-noised'' representation $\mathbf{\tilde y}_t$ is adopted to replace the quantized representation $\mathbf{\bar y}_t = \lfloor \mathbf{y}_t \rceil$ during model training, i.e., $\mathbf{\tilde y}_t = \mathbf{y}_t + \mathbf{o}$ with $o_i \sim \mathcal{U}(-\frac{1}{2},\frac{1}{2})$. Each ${\tilde y}_{t,i}$ is variationally modeled as a Laplace distribution with learned parameters ${\tilde \mu}_{t,i}$ and ${\tilde \sigma}_{t,i}$ such that
\begin{equation}\label{eq_prob_y}
\begin{aligned}
& p_{\mathbf{\tilde y}_t | \mathbf{\check c}_t, \mathbf{\tilde z}_t} (\mathbf{\tilde y}_t | \mathbf{\check c}_t, \mathbf{\tilde z}_t) = \prod_i p_{{\tilde y}_{t,i} | \mathbf{\check c}_t, \mathbf{\tilde z}_t, \mathbf{\tilde y}_{t,<i}} ({\tilde y}_{t,i} | \mathbf{\check c}_t, \mathbf{\tilde z}_t, \mathbf{\tilde y}_{t,<i})\\
& = \prod_i \Big( {\mathcal{L}}({\tilde{\mu}}_{t,i},{\tilde{\sigma}}_{t,i}) * \mathcal{U}(-\frac{1}{2},\frac{1}{2}) \Big) ({\tilde y}_{t,i})\\
& \text{with~} ({\tilde \mu}_{t,i},{\tilde \sigma}_{t,i}) = {h_{{\rm{pf}}}}({h_{{\rm{hpd}}}}({{\mathbf{\tilde z}}_t}),{h_{{\rm{ar}}}}({{\mathbf{\tilde y}}_{t, < i}}),{h_{{\rm{tpe}}}}({{\mathbf{\check c}}_t})),
\end{aligned}
\end{equation}
where ``$*$'' is the convolutional operation, $i \in \{1,2,\dots,I\}$ denotes the path embedding index, and the proxy hyperprior ${{\mathbf{\tilde z}}_t}$ is obtained by performing the hyperprior encoder network $h_{\rm{hpe}}(\cdot)$ on $\mathbf{y}_t$ and adding the uniformly sampled random offset $\mathbf{o}$, i.e., ${{\mathbf{\tilde z}}_t} = h_{\rm{hpe}}(\mathbf{y}_t) + \mathbf{o}$. Since we do not have prior beliefs about the hyperprior $\mathbf{\tilde z}_t$, it can be modeled as non-parametric fully factorized density \cite{balle2018}, i.e.,
\begin{equation}\label{eq_entropy_model_z}
  p_{\mathbf{\tilde z}_t} (\mathbf{\tilde z}_t) = \prod_j \left( p_{{z}_{t,j} | \bm{\psi}^{(j)}} ({z}_{t,j} | \bm{\psi}^{(j)}) * \mathcal{U}(-\frac{1}{2},\frac{1}{2}) \right) ({\tilde z}_{t,j}),
\end{equation}
where $\bm{\psi}^{(j)}$ encapsulates all the parameters of $p_{{z}_{t,j} | \bm{\psi}^{(j)}}$. During model testing, the entropy model $P_{\mathbf{\bar z}_t}$ is established by taking discrete values from the learned entropy model $p_{\mathbf{\tilde z}_t}$ by substituting $\mathbf{\tilde z}_t$ as $\mathbf{\bar z}_t$. $h_{\rm{hpd}}(\cdot)$ denotes the hyperprior decoder network to provide the hierarchical prior. $h_{\rm{ar}}(\cdot)$ represents the auto regressive network to provide the spatial prior. $h_{\rm{tpe}}(\cdot)$ denotes the temporal prior encoder network to provide additional side information. $h_{\rm{pf}}(\cdot)$ denotes the prior fusion network operating on the above three types of prior information. During model testing, the entropy model $P_{{\bar y}_{t,i} | \mathbf{\check c}_t, \mathbf{\bar z}_t, \mathbf{\bar y}_{t,<i} }$ in \eqref{eq_entropy_yi} will be established by taking discrete values from the learned entropy model $p_{{\tilde y}_{t,i} | \mathbf{\check c}_t, \mathbf{\tilde z}_t, \mathbf{\tilde y}_{t,<i}}$ by substituting $\mathbf{\tilde y}_t$ and $\mathbf{\tilde z}_t$ as $\mathbf{\bar y}_t$ and $\mathbf{\bar z}_t$, respectively.

As stated in NTC \cite{balle2020nonlinear}, the probabilistic model of $\mathbf{y}_{t,i}$ can be conditioned on some other vector $\mathbf{\tilde{z}}_t$ like \cite{balle2018} or its preceding dimensions $\mathbf{y}_{t,<i}$ as that in \cite{minnen2018}. The former corresponds to forward adaptation (FA) of the density model, and the latter is backward adaptation (BA). In this paper, due to the auto-regressive computation in \eqref{eq_prob_y}, our entropy model is established under the BA mode. One can also use the FA mode, where $h_{\rm{pf}}(\cdot)$ relies only on the hierarchical prior and the context, i.e., $({\tilde \mu}_{t,i},{\tilde \sigma}_{t,i}) = {h_{{\rm{pf}}}}({h_{{\rm{hpd}}}}({{\mathbf{\tilde z}}_t}),{h_{{\rm{tpe}}}}({{\mathbf{\check c}}_t}))$. Herein, the FA mode in DVST however cannot harvest the gain of better codec parallelism while it incurs performance degradation. The reason is that the auto-regressive computations of BA in our DVST are only used in entropy modeling to estimate the probability. The following deep JSCC codec runs in parallel for each embedding $y_{t,i}$. In comparison, traditional codec relying on arithmetic coding also adopts regressive computations in the BA mode which leads to higher latency. Therefore, we advertise the BA mode in our DVST framework.

With the learned entropy model $P_{{\bar y}_{t,i} | \mathbf{\check c}_t, \mathbf{\bar z}_t, \mathbf{\bar y}_{t,<i} }$, the allocated channel bandwidth cost $k_{t,i}$ to transmit the embedding $y_{t,i}$ is formulated as
\begin{equation}\label{eq_rate_allocation}
k_{t,i} = {\eta_{t}}{r_{t,i}} = -{\eta_{t}}\log{P_{{\bar y}_{t,i} | \mathbf{\check c}_t, \mathbf{\bar z}_t, \mathbf{\bar y}_{t,<i} }({\bar y}_{t,i} | \mathbf{\check c}_t, \mathbf{\bar z}_t, \mathbf{\bar y}_{t,<i} )},
\end{equation}
where the scaling factor ${\eta_{t}}$ denotes the proportion from the entropy of embedding $y_{t,i}$ to the number of channel symbols. In particular, the physical meaning of ${\eta_{t}}$ can be interpreted as ${\eta_{t}} = {\eta_{t}^{\prime} / C}$, where $C$ is the channel capacity (bits per channel symbol), and $\eta_{t}^{\prime}$ denotes an efficiency factor representing the capability of deep JSCC codec. Accordingly, $\eta_{t}^{\prime} = 1$ stands for the ideal JSCC codec that is of the same performance as entropy-achieving source coding combined with capacity-achieving channel coding.

Following the aforementioned entropy model, the channel bandwidth cost of the primary link for transmitting semantic features $\mathbf{y}_t$ is derived as
\begin{equation}\label{eq_channel_cost_primary}
\begin{aligned}
k_t^{\rm{pl}} & = \sum\limits_i{k_{t,i}} \\
~ & = \sum\limits_i{-{\eta_{t,i}}\log{P_{{\bar y}_{t,i} | \mathbf{\check c}_t, \mathbf{\bar z}_t, \mathbf{\bar y}_{t,<i} }({\bar y}_{t,i} | \mathbf{\check c}_t, \mathbf{\bar z}_t, \mathbf{\bar y}_{t,<i} )}}.
\end{aligned}
\end{equation}

\subsection{Motion Transmission and Context Learning}

As for context learning functions $\varphi_a(\cdot)$, $\varphi_s(\cdot)$, $\gamma_e(\cdot)$, and $\gamma_d(\cdot)$, inspired by \cite{DCVC}, we also adopt the idea of motion estimation and motion compensation (MEMC) to formulate specific forms of these functions. Different from conventional MEMC applied in source pixel domain, we perform MEMC in semantic feature domain to generate contexts $\mathbf{\check c}_t$, $\mathbf{\hat c}_t$ and in deep JSCC codeword domain to generate contexts $\mathbf{\check d}_t$, $\mathbf{\hat d}_t$. This paradigm utilizes rich information density in feature/codeword domain to fascinate high-efficiency video transmission over wireless channels with limited bandwidth.

As shown in Fig. \ref{Fig3}, the motion vector (MV) $\mathbf{m}_t$ is generated by using the flow estimation network \cite{ranjan2017optical} performed between the current frame $\mathbf{x}_t$ and the reconstructed reference frame $\mathbf{\check x}_{t-1}$. This MV is then transmitted over the wireless channel, which is referred to as the \emph{motion link}. The whole process copies from the primary link without using context, i.e.,
\begin{equation}\label{eq_motion_trans_process}
\begin{aligned}
{{\mathbf{m}}_t} \xrightarrow{{{g_a^{\rm{mv}}}( \cdot)}} {{\mathbf{y}}_t^{\rm{mv}}}
\xrightarrow{{{f_e^{\rm{mv}}}( \cdot)}} {{\mathbf{s}}_t^{\rm{mv}}}
\xrightarrow{{{W}( \cdot |\bm{\nu}})} {{\mathbf{\hat s}}_t^{\rm{mv}}}\\
 \xrightarrow{{{f_d^{\rm{mv}}}( \cdot)}} {{\mathbf{\hat y}}_t^{\rm{mv}}}
\xrightarrow{{{g_s^{\rm{mv}}}( \cdot)}} {{\mathbf{\hat m}}_t}.
\end{aligned}
\end{equation}
In our DVST design, we use ANN to realize each function in \eqref{eq_motion_trans_process} except for the channel transfer function $W$. The reference MV ${{\mathbf{\check m}}_t}$ used at the transmitter is generated by simulating the motion link without passing over the wireless channel, i.e.,
\begin{equation}\label{eq_motion_trans_process_simulate}
{{\mathbf{s}}_{t}^{\rm{mv}}} \xrightarrow{{{f_d^{\rm{mv}}}( \cdot)}} {{\mathbf{\check y}}_t^{\rm{mv}}}
\xrightarrow{{{g_s^{\rm{mv}}}( \cdot)}} {{\mathbf{\check m}}_t}.
\end{equation}
where the codeword ${{\mathbf{s}}_{t}^{\rm{mv}}}$ is obtained from \eqref{eq_motion_trans_process}.

In analogue to the primary link, by using the learned entropy model $P_{{\bar y}_{t,j}^{\rm{mv}} | \mathbf{\bar z}_t^{\rm{mv}}, \mathbf{\bar y}_{t,<j}^{\rm{mv}} }$ on ${\mathbf y}_{t}^{\rm{mv}}$, the allocated channel bandwidth cost $k_{t,j}^{\rm{mv}}$ to transmit the embedding $y_{t,j}^{\rm{mv}}$ is formulated as
\begin{equation}\label{eq_rate_allocation}
k_{t,j}^{\rm{mv}} = {\eta_{t,j}^{\rm{mv}}}{r_{t,j}^{\rm{mv}}} = -{\eta_{t,j}^{\rm{mv}}}\log{P_{{\bar y}_{t,j}^{\rm{mv}} | \mathbf{\bar z}_t^{\rm{mv}}, \mathbf{\bar y}_{t,<j}^{\rm{mv}} }({\bar y}_{t,j}^{\rm{mv}} | \mathbf{\bar z}_t^{\rm{mv}}, \mathbf{\bar y}_{t,<j}^{\rm{mv}})}.
\end{equation}
The channel bandwidth cost of the motion link for transmitting MV semantic features ${\mathbf y}_{t}^{\rm{mv}}$ is derived as
\begin{equation}\label{eq_channel_cost_motion}
k_t^{\rm{ml}} = -\sum\limits_j{ {\eta_{t,j}^{\rm{mv}}}\log{P_{{\bar y}_{t,j}^{\rm{mv}} | \mathbf{\bar z}_t^{\rm{mv}}, \mathbf{\bar y}_{t,<j}^{\rm{mv}} }({\bar y}_{t,j}^{\rm{mv}} | \mathbf{\bar z}_t^{\rm{mv}}, \mathbf{\bar y}_{t,<j}^{\rm{mv}})}}.
\end{equation}
The hyperprior $\mathbf{z}_t^{\rm{mv}}$ is obtained as $\mathbf{z}_t^{\rm{mv}} = h_{\rm{hpe}}^{\rm{mv}}(\mathbf{y}_t^{\rm{mv}})$, and the entropy model $P_{\mathbf{\bar z}_t^{\rm{mv}}}$ of the quantized hyperprior $\mathbf{\bar z}_t^{\rm{mv}}$ can be obtained similar to \eqref{eq_entropy_model_z}.

\begin{figure*}[t]
\setlength{\abovecaptionskip}{0.cm}
\setlength{\belowcaptionskip}{-0.cm}
\centering{\includegraphics[scale=0.35]{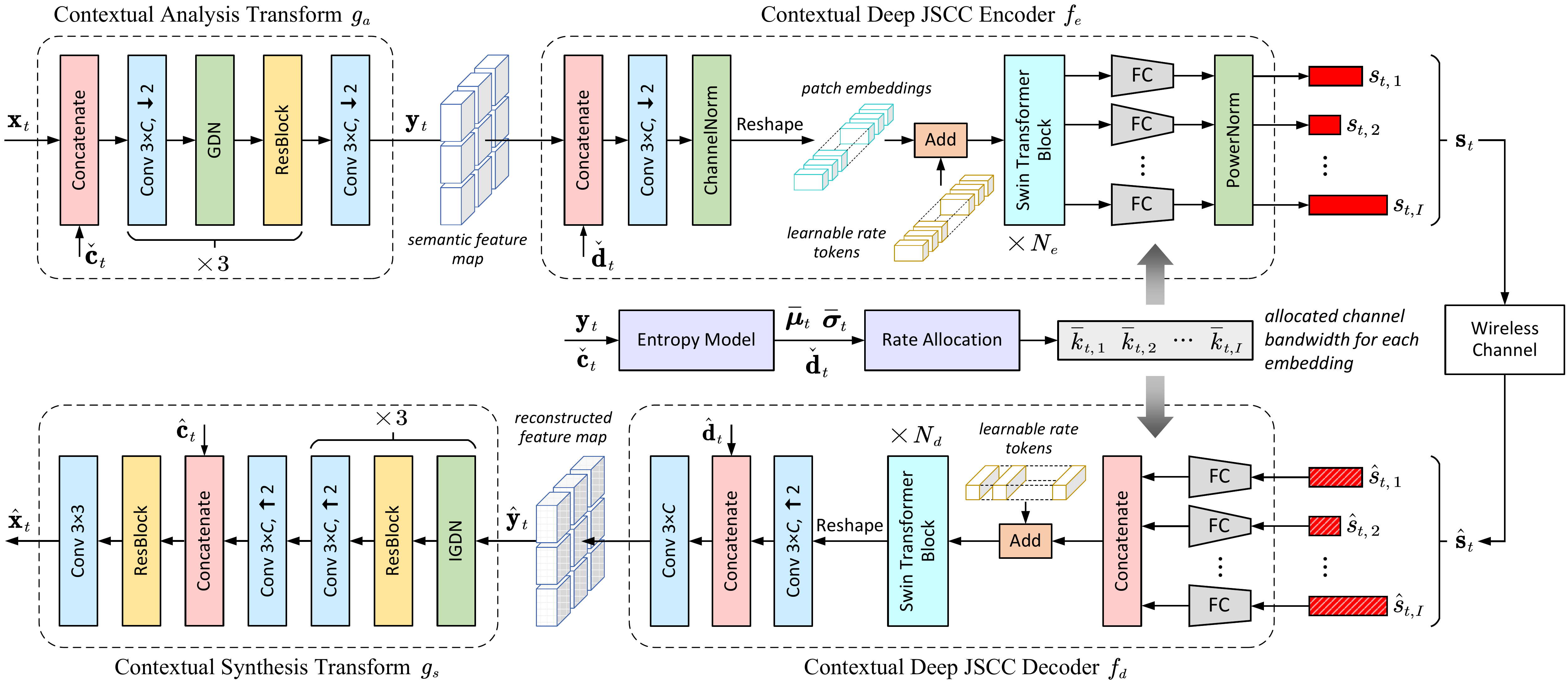}}
\caption{Network architectures of primary link. Conv $k \times C$ is a convolution with $C$ channels and $k \times k$ filters, and the followed $\uparrow 2 / \downarrow 2$ indicates upscaling or downscaling with a stride of $2$. ChannelNorm refers to a normalization layer from \cite{HIFIC}. GDN denotes the Generalised Divisive Normalization proposed in \cite{GDN}, and IGDN is the inverse GDN. FC denotes fully-connected layer.}\label{Fig4}
\vspace{0em}
\end{figure*}

As illustrated in Fig. \ref{Fig3}, at the transmitting end, the feature domain context generation function $\varphi_a$ is formulated as
\begin{equation}\label{eq_context_a}
{\mathbf{\check c}}_t = \varphi _a(\mathbf{\check{x}}_{t-1})=\varphi _{\mathrm{ref}}(\mathrm{warp(\varphi}_{\mathrm{fe}}(\mathbf{\check{x}}_{t-1}),\mathbf{\check{m}}_t)),
\end{equation}
where the reference frame $\mathbf{\check{x}}_{t-1}$ is obtained as \eqref{eq_video_trans_process_simulate}. We use the feature extractor network ${\varphi}_{\mathrm{fe}}(\cdot)$ to convert the reference frame $\mathbf{\check{x}}_{t-1}$ to its feature domain representation.
The reference MV guides where to extract feature domain context by using the warping function $\rm{warp}(\cdot)$ \cite{NIPS2015_warp}.
The refinement network $\varphi _{\mathrm{ref}}(\cdot)$ is used to restore the spatial discontinuity problem caused by warping operation.

As for the codeword context ${\mathbf{\check d}}_t$, its generation function $\gamma_e$ is formulated as
\begin{equation}\label{eq_context_e}
{\mathbf{\check d}}_t = \gamma_e(\mathbf{\check{y}}_{t-1})=\gamma _{\mathrm{ref}}(\mathrm{warp(\gamma}_{\mathrm{pc}}(\mathbf{\check{y}}_{t-1}),\mathbf{\check{m}}_t)).
\end{equation}
The precoding network ${\gamma}_{\mathrm{pc}}(\cdot)$ operates on the reference semantic feature map $\mathbf{\check{y}}_{t-1}$ which preprocesses the data before warping it. ${\gamma}_{\mathrm{ref}}(\cdot)$ denotes the refinement network.

At the receiving end, the context learning procedures are similar to \eqref{eq_context_a} and \eqref{eq_context_e}, where the realistic received MV $\mathbf{\hat m}_t$ is used. In addition, $\mathbf{\check{x}}_{t-1}$ and $\mathbf{\check{y}}_{t-1}$ are substituted by $\mathbf{\hat{x}}_{t-1}$ and $\mathbf{\hat{y}}_{t-1}$, respectively.

\subsection{Optimization Goal}\label{subsection_opt_goal}

The optimization goal of our DVST system is to use the least channel bandwidth cost to get the best video reconstruction quality or downstream task performance.
Given the reference frame, the loss function at the current time step $t$ is formulated as a rate-distortion (RD) form, i.e.,
\begin{equation}\label{eq_training_loss}
L_t = \lambda \cdot k_t + D_t = \lambda(k_t^{\rm{pl}} + k_t^{\rm{ml}}) + D_t,
\end{equation}
where $\lambda$ controls the trade-off between the total channel bandwidth cost $k_t$ and the distortion $D_t$.

Due to the conditional coding architecture, the performance of a previous frame will affect its subsequent frame. Therefore, during the training phase of DVST, we take into account the correlations of frames within a GOP. The DVST model can thus learn to allocate channel bandwidth resources within one frame and among various frames. Thus, the overall training loss function is formulated as
\begin{equation}\label{eq_GOP_training_loss}
L = \frac{1}{N}\sum_{t=1}^{N} \left( \lambda \cdot k_t + D_t \right) = \frac{1}{N}\sum_{t=1}^{N} \left( \lambda(k_t^{\rm{pl}} + k_t^{\rm{ml}}) + D_t \right).
\end{equation}
The training procedure details will be introduced in the subsequent subsection.

For machine-type semantic communications, our DVST can directly drive the downstream machine vision tasks while preserving the advantages of signal level reconstruction. Different from the functional transmission mode adopted in \cite{jankowski2020wireless}, this paper aims to transmit videos friendly to both human vision and machine analytics \cite{duan2020video}.
Therefore, we incorporate the low-level signal distortion and the loss of high-level tasks, thus, the distortion term is reformulated as $D_t = D_{t,\rm{rec}} + \beta D_{t,\rm{task}}$, where $D_{t,\rm{rec}}$ denotes the reconstruction loss and $D_{t,\rm{task}}$ denotes the loss of downstream task.

\section{Architectures and Implementations}\label{section_architecture}

In this section, we present details of the adopted network architectures to implement our DVST. Then, we introduce the progressive training strategy to enable a stable model learning.

\subsection{Network Architectures}

We illustrate ANN implementation details of the primary link in Fig. \ref{Fig4}, including the contextual nonlinear transform modules and the contextual deep JSCC modules. For brevity, we do not repeat the structure of the motion link since it has almost the same architecture as the primary link except that the MV is of two channels, and the contextual operations are removed.

\begin{figure*}[b]
\hrulefill
\begin{equation}\label{eq_ntc_mv_training_loss}
\begin{aligned}
  {\ddot{L}}_{t}^{\rm{mv}} & = \lambda  {\ddot{R}}_{t}^{\rm{mv}} +  {\ddot{D}}_{t}^{\rm{mv}} = \lambda \Big(
 -\sum_{j}\log{P_{{\tilde y}_{t,j}^{\rm{mv}} | \mathbf{\tilde z}_t^{\rm{mv}}, \mathbf{\tilde y}_{t,<j}^{\rm{mv}} }({\tilde y}_{t,j}^{\rm{mv}} | \mathbf{\tilde z}_t^{\rm{mv}}, \mathbf{\tilde y}_{t,<j}^{\rm{mv}})}
 -\log{P_{\mathbf{\tilde{z}}_t^{\rm{mv}}}(\mathbf{\tilde{z}}_t^{\rm{mv}})} \Big) + D\left( {\rm warp}(\mathbf{x}_{t-1},\mathbf{ \ddot{m}}_t), \mathbf{x}_t\right)  \\
  ~ & \text{with~} {\mathbf{\tilde y}}_t^{\rm{mv}} = \mathbf{ y}_t^{\rm{mv}} + \mathbf{ o},~{\mathbf{\tilde z}}_t^{\rm{mv}} = \mathbf{ z}_t^{\rm{mv}} + \mathbf{ o},~ \mathbf{y}_t^{\rm{mv}} = g_a^{\rm{mv}}(\mathbf{ m}_t),~\mathbf{z}_t^{\rm{mv}} = h_{\rm{hpe}}^{\rm{mv}}(\mathbf{y}_t^{\rm{mv}}),~\mathbf{\ddot{m}}_t = g_s^{\rm{mv}}(\mathbf{\tilde{y}}_t^{\rm{mv}}).
\end{aligned}
\end{equation}
\end{figure*}

\begin{figure*}[b]
    \hrulefill
    \begin{equation}\label{eq_ntc_res_training_loss}
    \begin{aligned}
    {\ddot{L}}_{t} & = \lambda {\ddot{R}}_{t} + {\ddot{D}}_{t} = \lambda \Big(
    -\sum_{j}\log{P_{{\tilde y}_{t,j} | \mathbf{\tilde z}_t, \mathbf{\tilde y}_{t,<j}}({\tilde y}_{t,j} | \mathbf{\tilde z}_t, \mathbf{\tilde y}_{t,<j})}
    -\log{P_{\mathbf{\tilde{z}}_t}(\mathbf{\tilde{z}}_t)} \Big)
    + D\left( \mathbf{ \ddot{x}}_t, \mathbf{x}_t\right)  \\
    ~ & \text{with~} {\mathbf{\tilde y}}_t = \mathbf{ y}_t + \mathbf{ o},~ {\mathbf{\tilde z}}_t = \mathbf{ z}_t + \mathbf{ o},~\mathbf{y}_t = g_a(\mathbf{ x}_t | \mathbf{\check{c}}_t),~\mathbf{z}_t = h_{\rm{hpe}}(\mathbf{y}_t),~\mathbf{\ddot{x}}_t = g_s(\mathbf{\tilde{y}}_t | \mathbf{\hat{c}}_t),
    \end{aligned}
    \end{equation}
\end{figure*}

\subsubsection{Contextual Nonlinear Transform}

\begin{figure}[t]
\setlength{\abovecaptionskip}{0.cm}
\setlength{\belowcaptionskip}{-0.cm}
\begin{center}
	\hspace{-.0in}
	\subfigure[]{
		\includegraphics[scale=0.33]{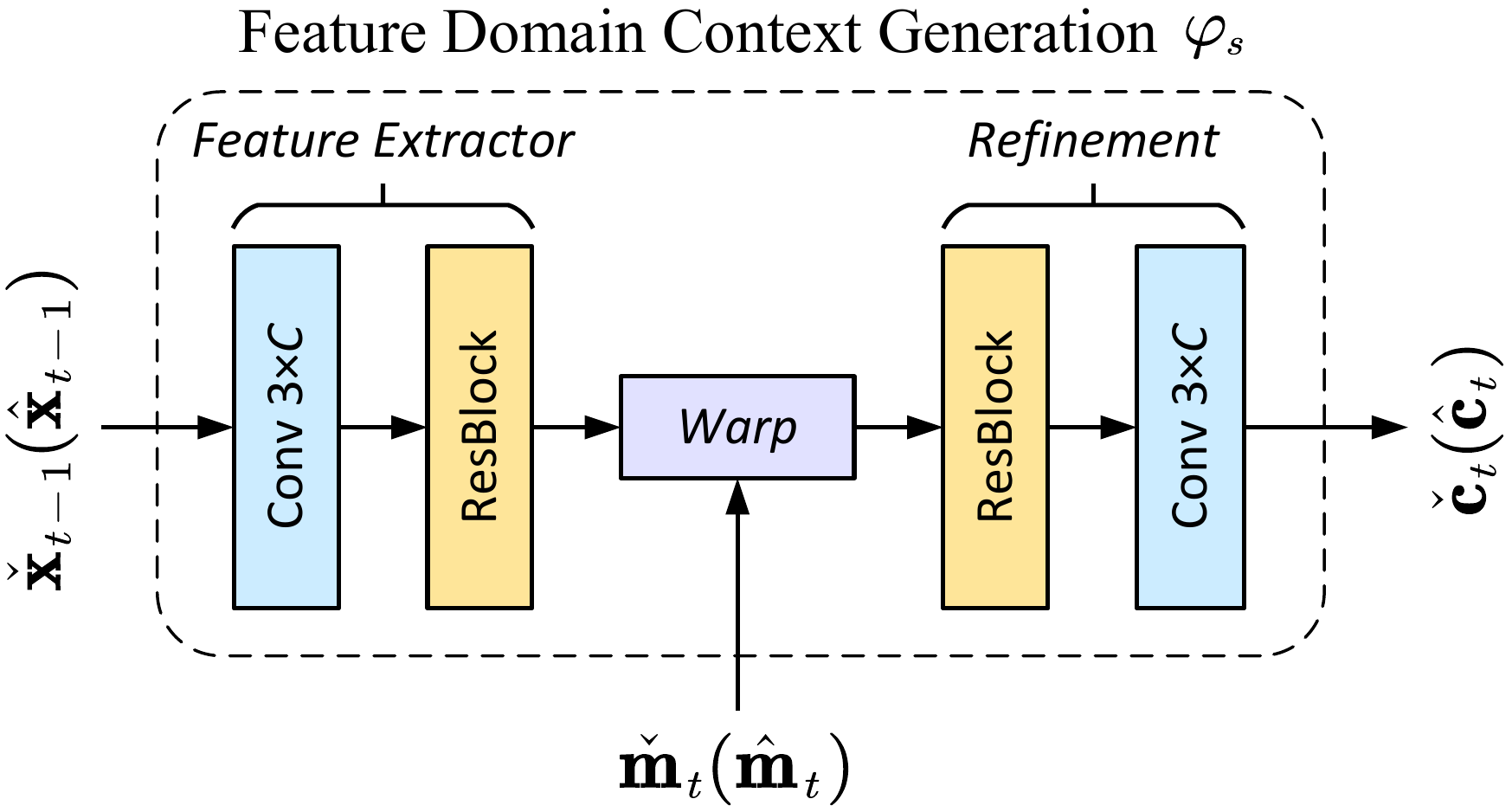}
	}
	\subfigure[]{
		\includegraphics[scale=0.33]{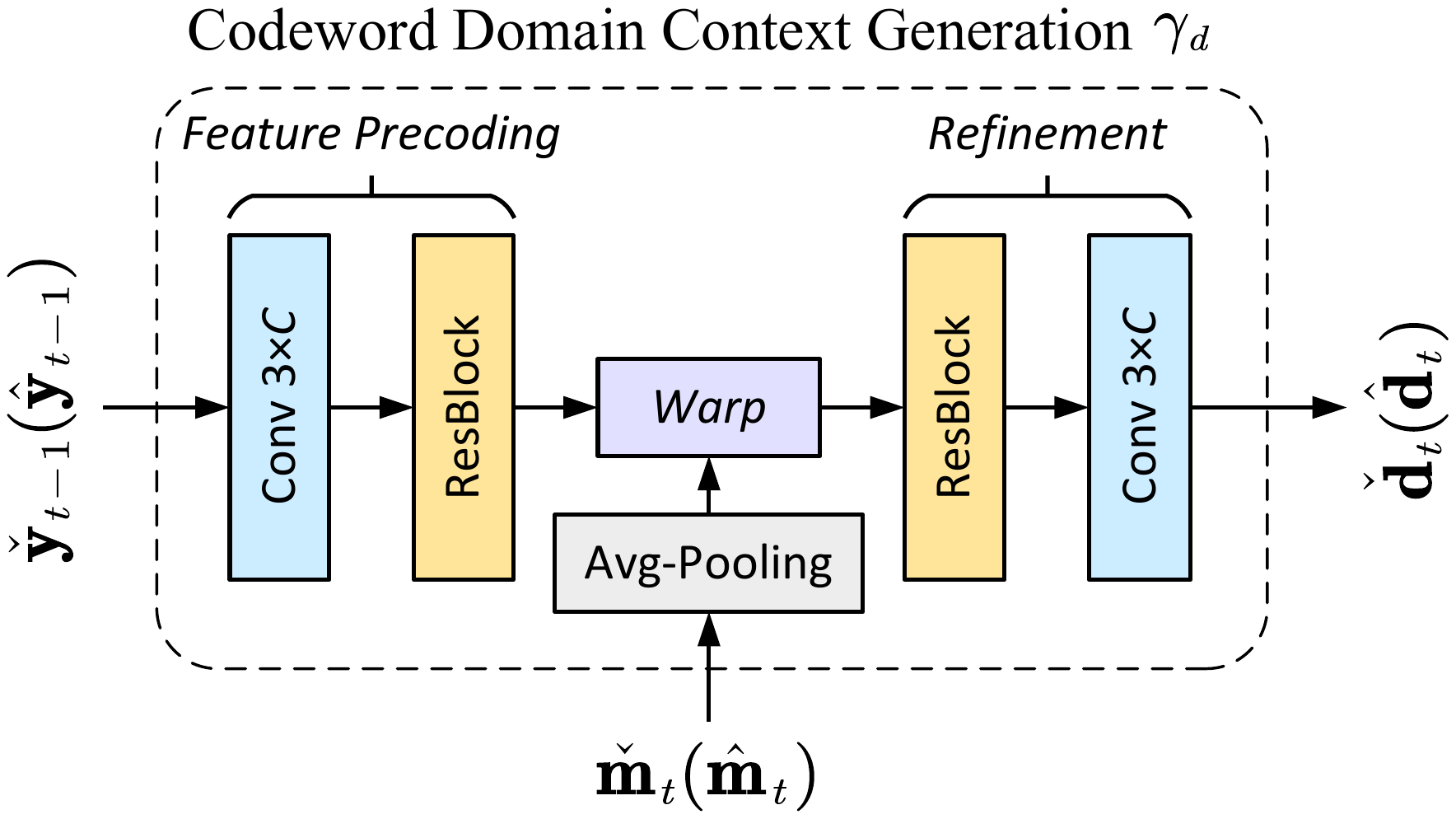}
	}
	\caption{Detailed structure of (a) semantic domain feature context generation networks, (b) deep JSCC codeword domain context generation networks.}
	\label{Fig5}
\end{center}
\end{figure}

The proposed DVST drives the current frame $\mathbf{x}_t$ in terms of the context extracted from MV rather than relying on a handcrafted subtraction operation. Fig. \ref{Fig5}(a) illustrates the context generation network $\varphi_a$ in semantic feature domain, which consists of feature extractor, warp, and refinement operation. After that, the analysis transform $g_a$ concatenates the Tx context $\mathbf{\check{c}}_t$ with the current frame $\mathbf{x}_t$ to learn a compact latent representation $\mathbf{y}_t$ in semantic feature domain. To estimate the spatial varying mean values and standard deviations of $\mathbf{y}_t$, we follow the hyperprior entropy model in \cite{DCVC}, which fuses hierarchical prior, spatial prior, and temporal prior. The contextual synthesis transform $g_s$ has a symmetric architecture with $g_a$ except that it uses the Rx context $\mathbf{\hat{c}}_t$ generated from $\varphi_s$ which is of the same structure with $\varphi_a$.

\subsubsection{Contextual Deep JSCC}

Contextual deep JSCC module exploits the codeword contexts extracted from $\mathbf{\check y}_{t-1}$ and $\mathbf{\hat y}_{t-1}$ to collaboratively transmit the current latent representation $\mathbf{y}_t$. It is of variable rates in accordance to the estimated entropy. In particular, the structure of codeword context generator $\gamma_e$ is shown in Fig. \ref{Fig5}(b). Before the wrapping operation \cite{NIPS2015_warp}, we align the MV with the precoded feature by using an average pooling operation with stride 16. By using the learned entropy model, we pre-allocate the channel bandwidth cost of each spatial position $k_{t,i}$ as \eqref{eq_channel_cost_primary}. Accordingly, the encoder $f_e$ fuses the $\mathbf{y}_t$ with the context $\mathbf{\check{d}}_t$ and partitions the fused feature map into patch embedding sequence $\left( y_{t,1}, y_{t,2}, \cdots, y_{t,I} \right) $, where each embedding is an $M$-dimensional feature vector. After that, the practical channel bandwidth cost ${\bar k}_{t,i}$ for transmitting $y_{t,i}$ is determined as ${\bar k}_{t,i} = Q(k_{t,i})$. Herein, $Q$ denotes a scalar quantization whose range includes $2^{q}$ ($q = 1,2,\dots$) integers, and the quantization value set $\mathcal{V} = \{ v_1, v_2, \dots, v_{2^{q}} \}$ is related to the scaling factor $\eta$ and the Lagrange multiplier $\lambda$ in the RD loss function. In this way, we inform the receiver which rate is allocated to each embedding $y_i$ by transmitting predetermined $q$ bits as extra side information.

Instead of naively training $2^{q}$ deep JSCC networks, we exploit the dynamic neural network structure to realize variable rate transmission. As shown in Fig. \ref{Fig4}, $f_e$ consists of a powerful shared backbone to extract the contextual dependencies among $y_{t,i}$, and light FC layers to encode $y_{t,i}$ into the given dimension ${\bar k}_{t,i}$. In particular, we employ a group of $2^{q}$ FC layers with  different output dimensions $\{ v_1, v_2, \dots, v_{2^{q}} \}$, and each FC layer is invoked on demand. As a result, during the model forward pass, some FC layers may not be used while others may be used more than once. Additionally, to enhance the capacity of deep JSCC and extract the global and long-term correlations, we employ Swin Transformer blocks as the network backbone \cite{liu2021Swin}. Inspired by the application of positional embeddings in vision Transformer, we develop a group of rate tokens $\mathcal{R} = \{ {r}_{v_1}, {r}_{v_2}, \dots, {r}_{v_{2^{k_q}}} \}$ to indicate the CBR information. The rate tokens can be viewed as learnable parameters within Transformer. As shown in Fig. \ref{Fig4}, each embedding $y_{t,i}$ will be added with its corresponding rate token ${r}_{{\bar k}_{t,i}}$ before fed into Transformer blocks.
Hence, the output patch embeddings can get a better trade-off between fidelity and robustness. As a result, the following FC layer can efficiently rescale the dimensions of channel-input symbols $s_{t,i}$.

\subsection{Progressive Training Strategy}

As aforementioned, the goal of DVST is to minimize a compromise between channel bandwidth cost (including primary link and motion link) and end-to-end distortion (reconstruction error or downstream task accuracy). Starting from a pretrained optical flow estimation network, the training procedure consists of the following steps:
\begin{enumerate}[(1)]
\item Pretrain the nonlinear transform components of the motion link, including the motion estimation network, $g_a^{\rm{mv}}$, $g_s^{\rm{mv}}$, and the entropy model. In this step, we use the lossless previous frame as reference. The pretraining loss function ${\ddot{L}}_{t}^{\rm{mv}}$ of this step is formulated as \eqref{eq_ntc_mv_training_loss}, where $\mathbf{o}$ denotes uniformly sampled random quantization offsets.

\item Taking the wireless transmission error of the motion link into account, based on the previous step, deep JSCC codec $f_e^{\rm{mv}}$, $f_d^{\rm{mv}}$, and the rate adaptation module are jointly trained with nonlinear transform components to execute an MV transmission task. We also add the pretraining distortion terms ${\ddot{D}}_{t}^{\rm{mv}}$ to make the training process stable. The training loss function of this step is formulated as
    \begin{equation}\label{eq_ml_training_loss}
    L_t^{\rm{ml}} = {\ddot{D}}_{t}^{\rm{mv}} + D\left( {\rm warp}(\mathbf{x}_{t-1},\mathbf{ \hat{m}}_t), \mathbf{x}_t\right) + \lambda {k}^{\rm{ml}}_t ,
    \end{equation}
    where the reconstructed MV $\mathbf{ \hat{m}}_t$ is obtained as \eqref{eq_motion_trans_process} by passing over wireless channel, and the channel bandwidth cost of the motion link is obtained from \eqref{eq_channel_cost_motion}.

\item Pretrain the nonlinear transform components of the primary link and context generation networks $\varphi_a$ and $\varphi_s$ meanwhile freezing the parameters of motion link. Similar to the pretraining process of motion link, we still employ the ideal lossless previous frame as reference, i.e., we manually set $\mathbf{\check{x}}_{t-1} = \mathbf{\hat{x}}_{t-1} = \mathbf{x}_{t-1}$. In practice, following \cite{DCVC}, we temporally remove the bitcost terms and readd them after several training epoches. This strategy helps model to generate useful contexts and exploit them, which actually accelerates the convergence. The training loss of this step is formulated as \eqref{eq_ntc_res_training_loss}, where the generation of feature domain context $\mathbf{\check{c}}_t$ and $\mathbf{\hat{c}}_t$ can refer to (\ref{eq_context_a}).
    \item Based on the previous step, train the whole framework except the freezing motion link. The training loss of this step is formulated as
    \begin{equation}\label{eq_pl_training_loss}
    L_t^{\rm{pl}} = {\ddot{D}}_{t} + D\left(\mathbf{\hat{x}}_t, \mathbf{x}_t\right) + \lambda k^{\rm{pl}}_t ,
    \end{equation}
    where the reconstructed frame $\mathbf{ \hat{x}}_t$ is derived from \eqref{eq_video_trans_process}, and the channel bandwidth cost of the primary link is obtained as \eqref{eq_channel_cost_primary}. In this step, model learns to transmit current frame efficiently with the help of context $\mathbf{d}_t$ in JSCC codeword space. Also, for stable training, the end-to-end distortion ${\ddot{D}}_{t}$ is added.

\item Unfreeze the motion link and train the whole DVST model according to subsection \ref{subsection_opt_goal}. The final training loss for a GOP is formulated as
\begin{equation}\label{eq_total_training_loss}
L = \frac{1}{N}\sum_{t=1}^{N} \left( \lambda (k^{\rm pl}_t + k^{\rm mv}_t) + D\left(\mathbf{\hat{x}}_t, \mathbf{x}_t\right) + D\left(\mathbf{\ddot{x}}_t, \mathbf{x}_t\right) \right),
\end{equation}
where the pretraining distortion $D\left(\mathbf{\ddot{x}}_t, \mathbf{x}_t\right)$ serves as a regularization term to improve the training stability. In this step, the model indeed learns a bandwidth cost trade-off between primary and motion links. Moreover, due to the integrated training within a whole GOP, the rate allocation between frames has also been optimized, which will be shown in the ablation study of the subsequent section.
\end{enumerate}

\section{Experimental Results}\label{section_results}

\subsection{Experimental Setup}

\subsubsection{Datasets}

Our DVST model is trained with the Vimeo-90k dataset \cite{vimeo90k}, which consists of 89800 video clips with a large variety of scenes and actions. During the model training, the chunks are randomly cropped to $256\times 256$ pixels. We use $N=7$ unrolled frames as a GOP in the last step of the training procedure and disallow the gradients passing from the I-frame reconstruction to the P-frame.
We evaluate the performance of DVST using the HEVC test dataset \cite{bossen2013common} and the UVG dataset. As widely used standards to measure video-related algorithms' performance, they contain sequences of various content, frame rate, and resolution. In particular, the HEVC dataset includes Class A ($2560\times 1600$), Class B ($1920\times 1080$), Class C ($832\times 480$), Class D ($416\times 240$), Class E ($1280\times 720$). And the UVG dataset consists of $7$ videos with the resolution of $1920\times 1080$. During model testing, we set the GOP size as $N=4$, which is identical to the end-to-end wireless video transmission scheme in \cite{tung2021deepwive}. As for I-frame coding, we adopt our previous work of image semantic transmission using nonlinear transform source-channel coding \cite{dai2021nonlinear}.

\subsubsection{Implementation Details}
	
In all experiments, the channel dimension $C$ in Fig. \ref{Fig4} is set to 96 for the primary link and 128 for the motion link. In addition, the channel dimension $C$ in Fig. \ref{Fig5} is 96. As mentioned before, we employ the Swin Transformer \cite{liu2021Swin} as the backbone of the contextual deep JSCC codec, which greatly reduces the computation complexity of vision Transformers by conducting multi-head self-attention (MHSA) within local windows or shifted windows. In this paper, the number of Swin Transformer blocks is set to $N_e = N_d = 4$, and we use 8 heads and $16\times 16$ window size in MHSA. In addition, the quantized channel bandwidth cost value set of primary link is chosen as $\mathcal{V}_{\rm pl} = \{ 0, 2, 4, 6, 8, 10, 15, 20, 26, 32, 40, 48, 56, 64, 80, 96\} $, and the motion link $\mathcal{V}_{\rm ml} = \{ 0, 1, 2, 4, 8, 16, 32, 48\} $. Hence, extra side information of total $q = 7$ bits will be transmitted to inform the receiver of the CBR for each embedding. Since we adopt a large patch size of $32 \times 32$ pixels, the side information cost is relatively trivial compared to video content. The composition of total CBR will be discussed in the ablation study.

For the reconstruction task, we optimize DVST in terms of the mean squared error (MSE) for the peak-signal-to-noise ratio (PSNR) metric, or multiscale structural similarity \cite{msssim} (MS-SSIM) for perceptual quality. Multiple DVST models are trained with $\lambda \in \{1/4, 1/8, 1/16, 1/32, 1/64\}$ for MS-SSIM and $\lambda \in \{256, 128, 64, 32, 16\}$ for PSNR, thus achieving different RD tradeoffs. A smaller value of $\lambda$ leads to a larger CBR. We denote these models as ``DVST (PSNR)'' and ``DVST (MS-SSIM)'', respectively. For each model, we use the Adam optimizer \cite{ADAM} with a learning rate of $10^{-4}$. We use a mini-batch size of 8, and it takes about one week to train the whole DVST model on single RTX 3090 GPU.

\subsubsection{Comparison Schemes}

\begin{figure*}[t]
\setlength{\abovecaptionskip}{0.cm}
\setlength{\belowcaptionskip}{-0.cm}
\begin{center}
\hspace{.02in}
\subfigure[]{\includegraphics[scale=0.35]{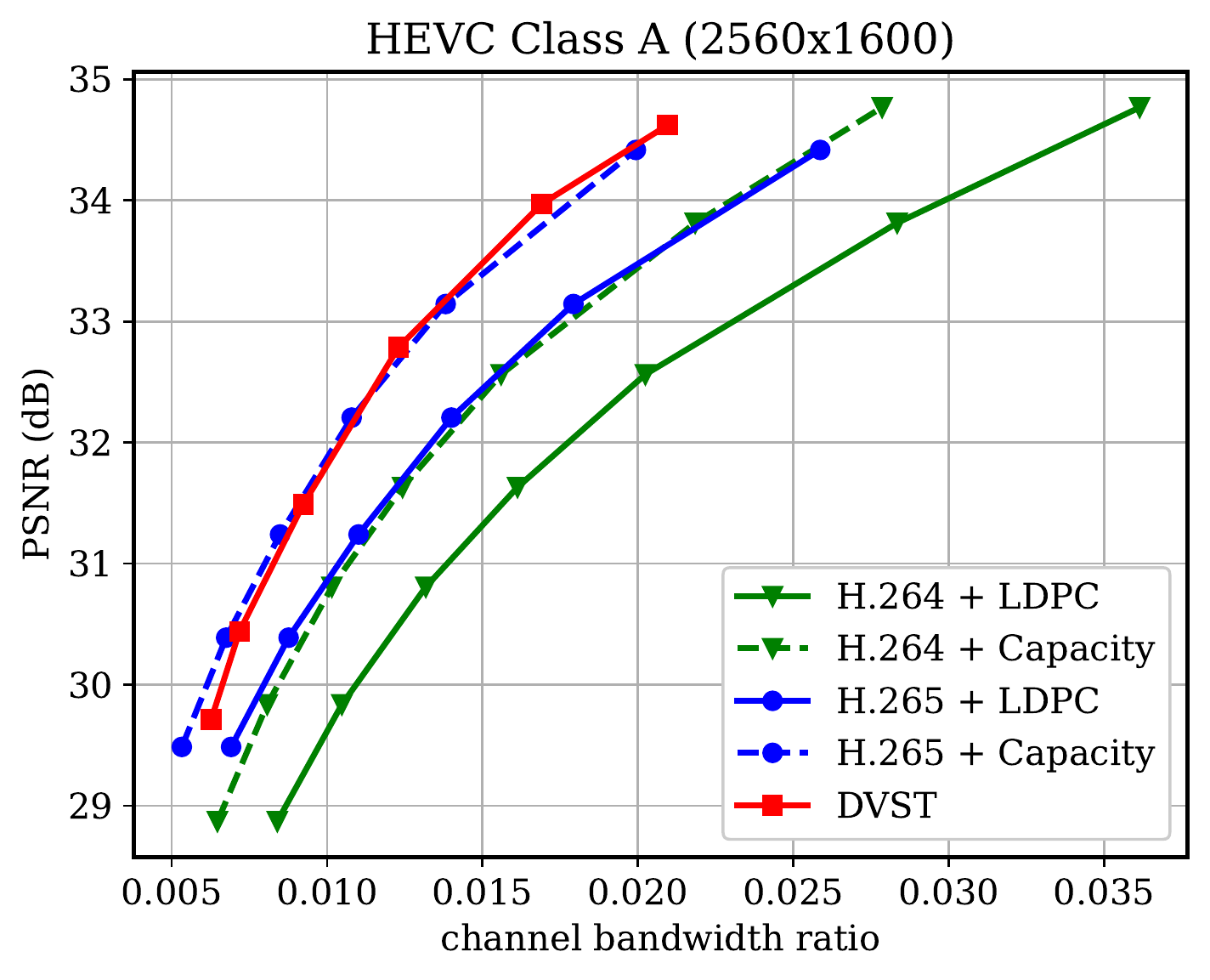}
}
%\hspace{-.1in}
\quad
\subfigure[]{
	\includegraphics[scale=0.35]{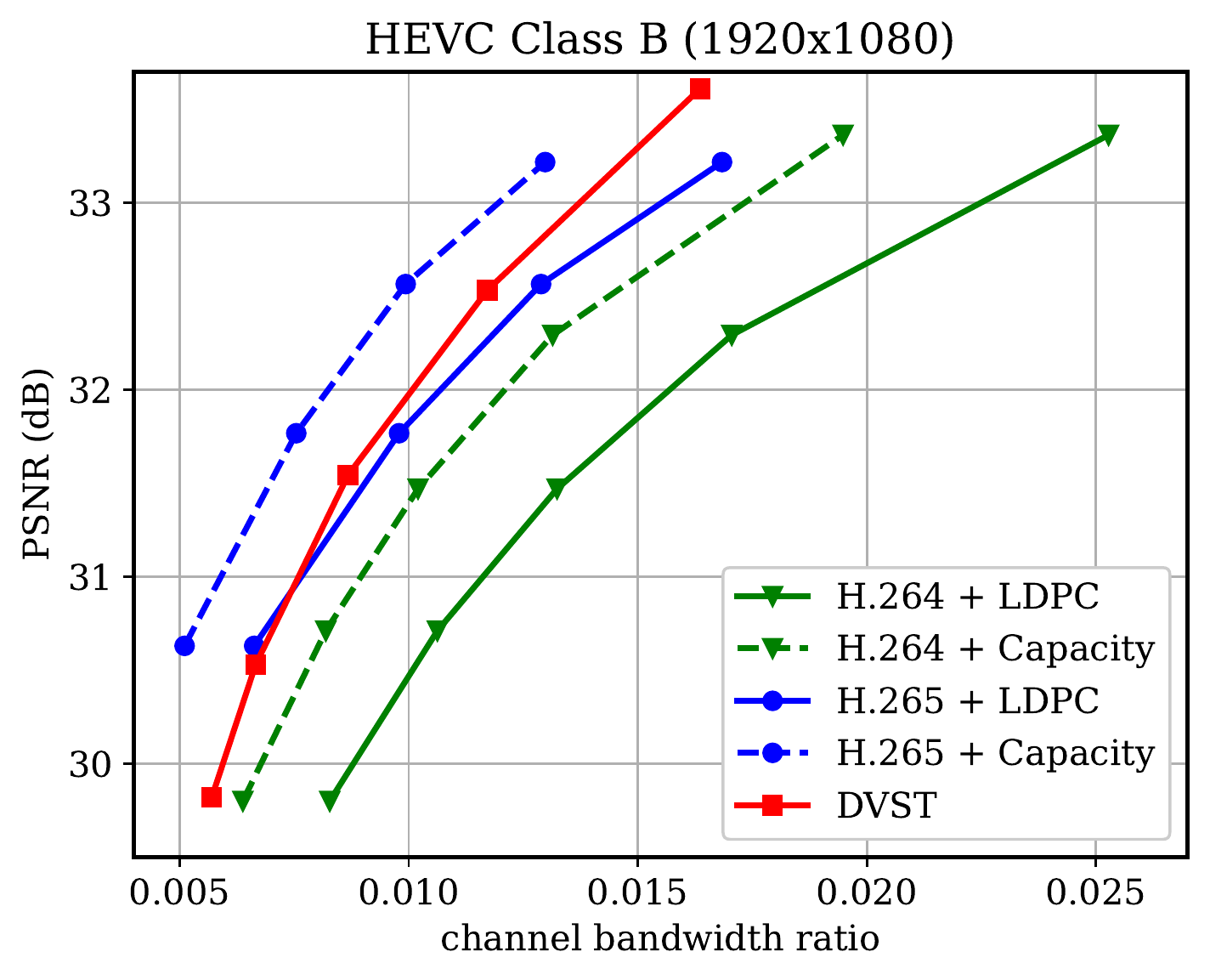}
}
\hspace{-.05in}
\quad
\subfigure[]{
	\includegraphics[scale=0.35]{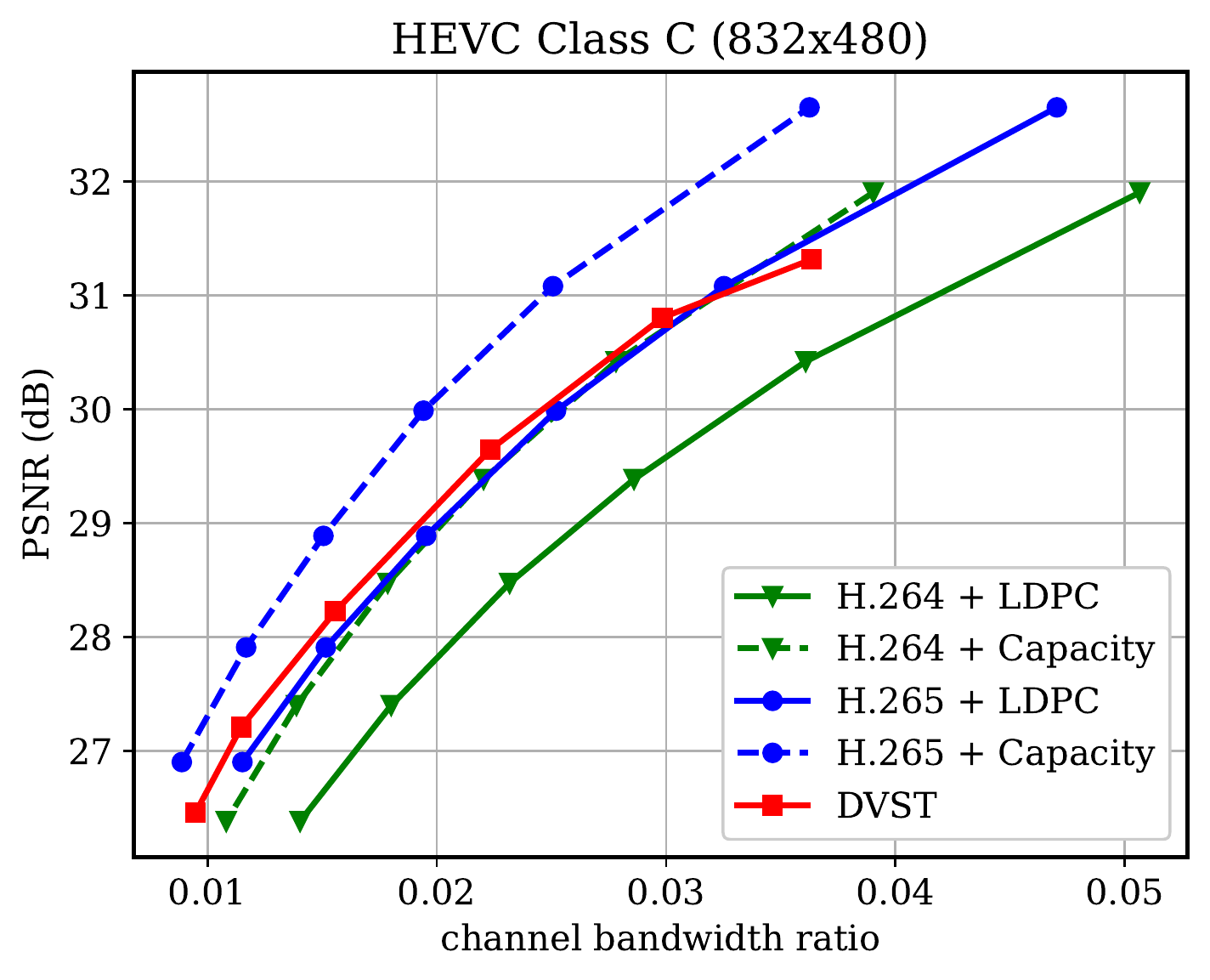}
}
%\hspace{-.1in}
\\
\subfigure[]{
	\includegraphics[scale=0.35]{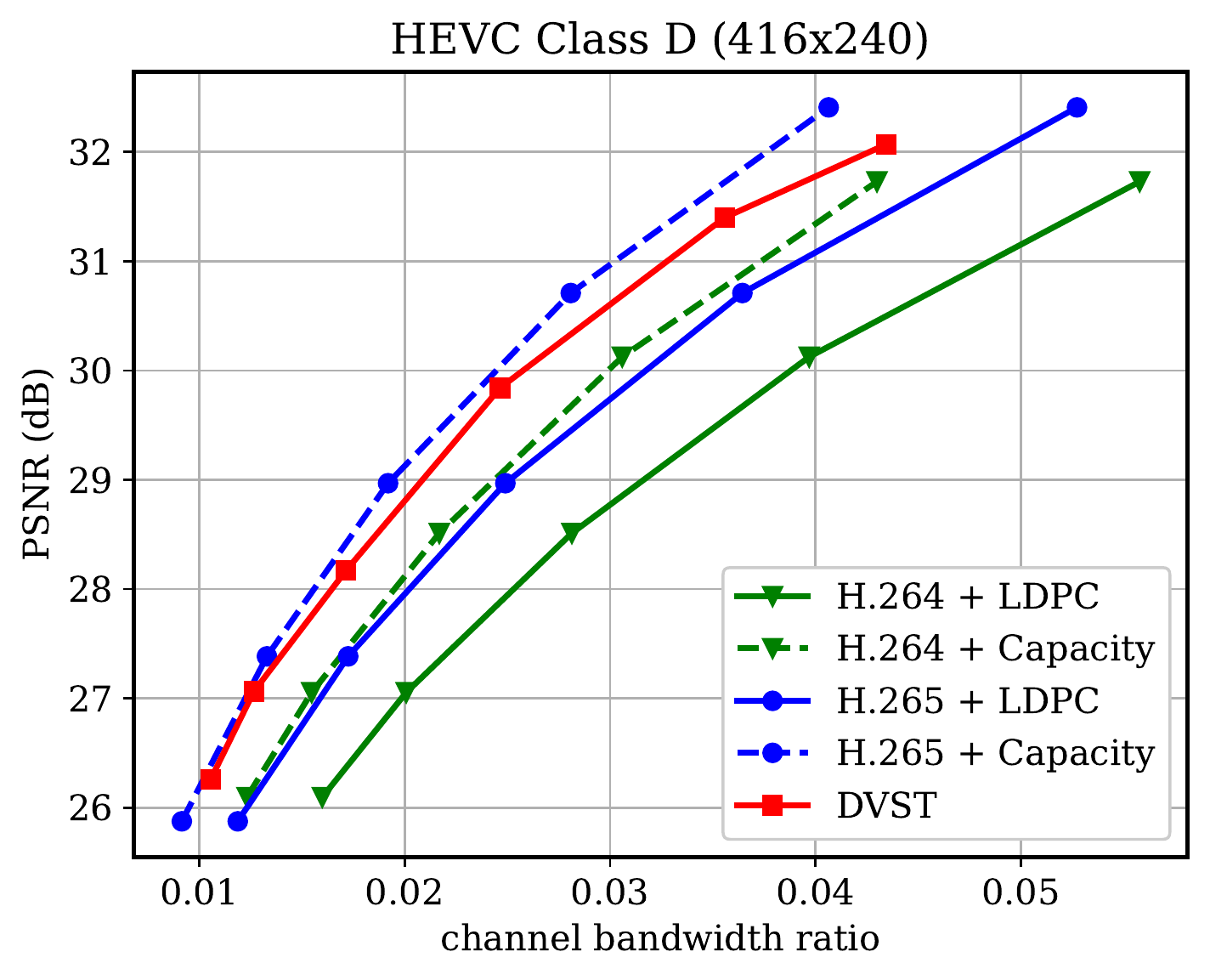}
}
%\hspace{-.1in}
\quad
\subfigure[]{
	\includegraphics[scale=0.35]{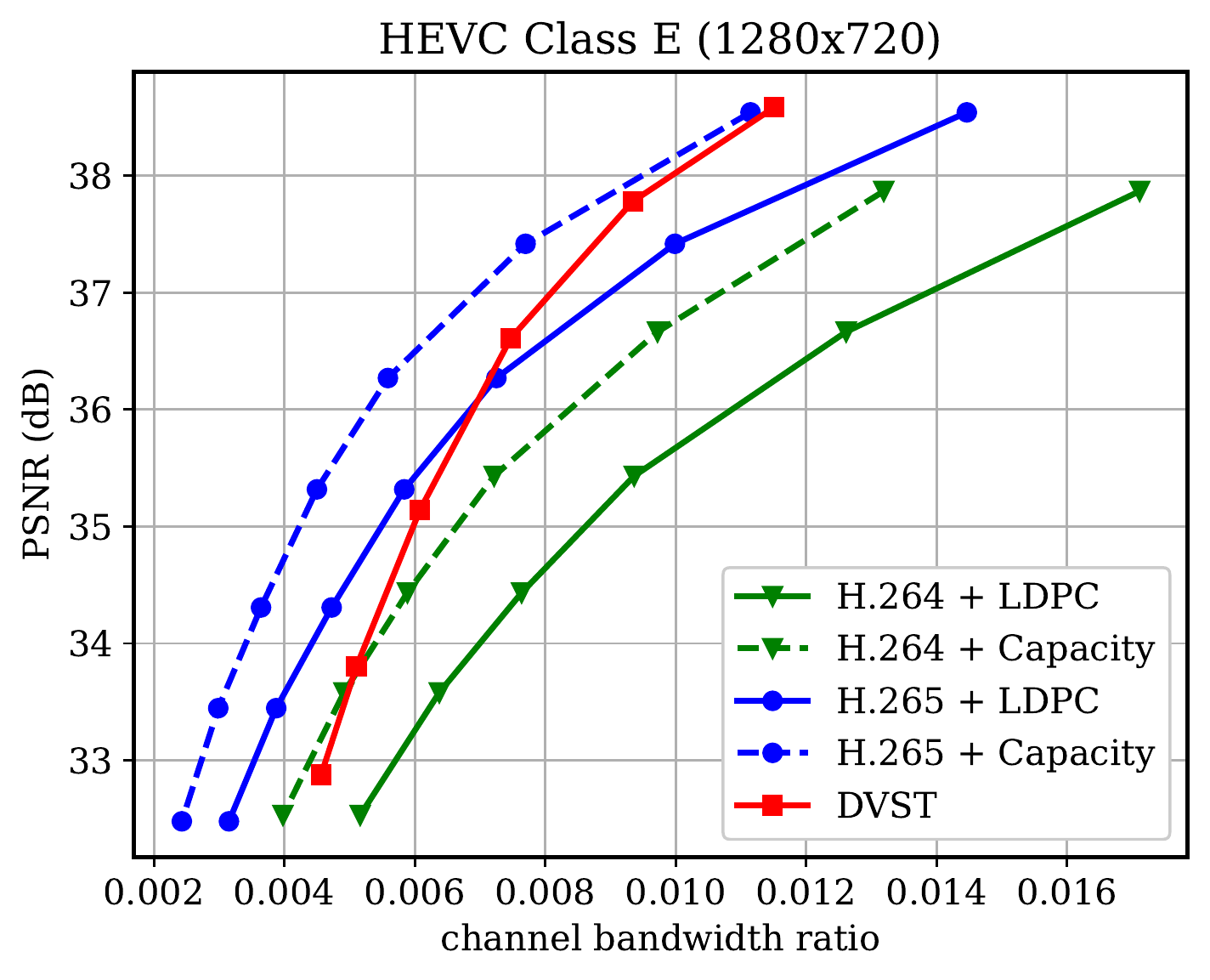}
}
%\hspace{-.1in}
\quad
\subfigure[]{
	\includegraphics[scale=0.35]{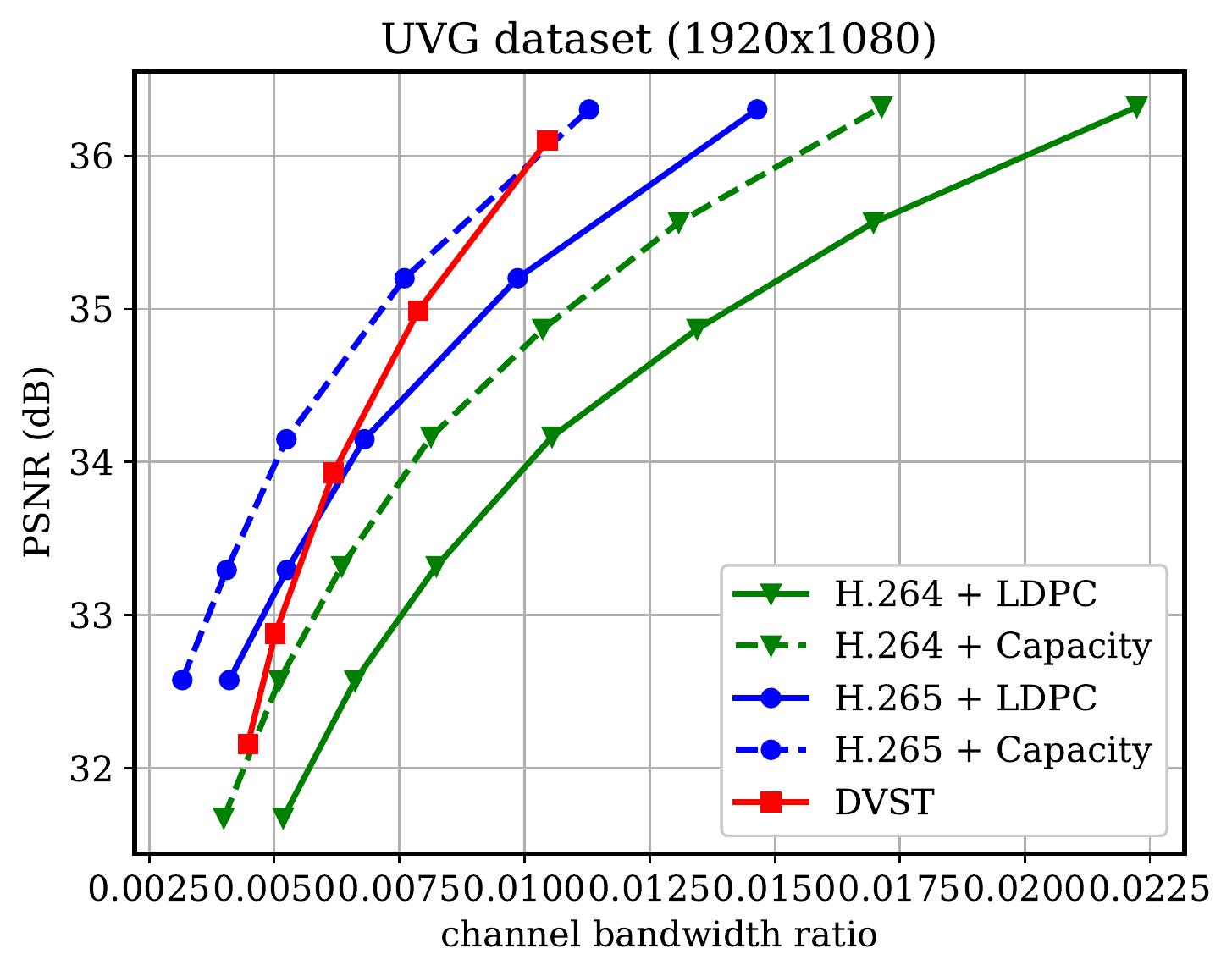}
}

\caption{PSNR performance versus the average channel bandwidth ratio (CBR) over the AWGN channel at $ \text{SNR} = 10$dB.}
\label{Fig6}
\end{center}
\end{figure*}

Following \cite{tung2021deepwive}, we compare our DVST with classical video coded transmission schemes in current mainstream wireless communication systems. In particular, we employ the standard video codecs (H.264 \cite{H264} and H.265 \cite{H265}) for source coding combined with practical LDPC codes \cite{richardson2018design} or ideal capacity-achieving channel code family for channel coding.
For brevity, we use ``+'' to concatenate the source coding and channel coding schemes, e.g., H.265 combined with capacity-achieving channel code is denoted as ``H.265 + Capacity''. As we shall note, the ideal ``H.264 + Capacity'' or ``H.265 + Capacity'' scheme can be viewed as a performance upper bound on traditional separation-based source and channel coding schemes. The above simulations are implemented on the top of Sionna \cite{sionna}, an open-source library for the link-level simulation of digital communication systems. In addition, we refer to the configurations of H.264 and H.265 in \cite{DVC}, which adopt the typical \emph{ffmpeg} settings for low-latency and veryfast mode. In practical implementation, to be aligned with previous works \cite{DJSCC}, we also convert two consecutive real symbols in $\mathbf{s}$ as one complex channel-input symbol and add complex Gaussian noise.

\subsection{Reconstruction Task Results}

\begin{figure*}[t]
	\setlength{\abovecaptionskip}{0.cm}
	\setlength{\belowcaptionskip}{-0.cm}
	\begin{center}
		\subfigure[]{\includegraphics[scale=0.35]{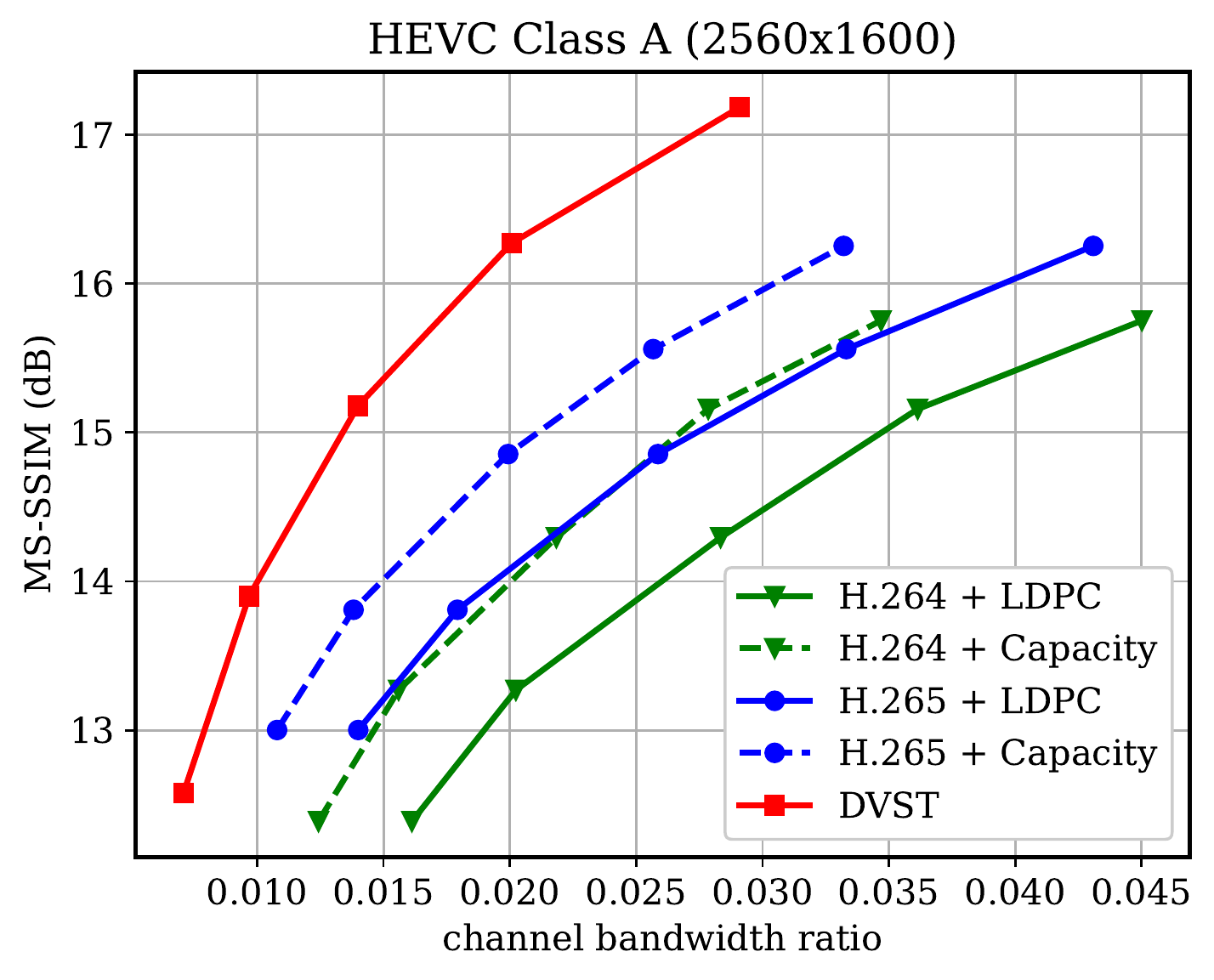}
		}
        \hspace{-.025in}
		\quad
		\subfigure[]{
			\includegraphics[scale=0.35]{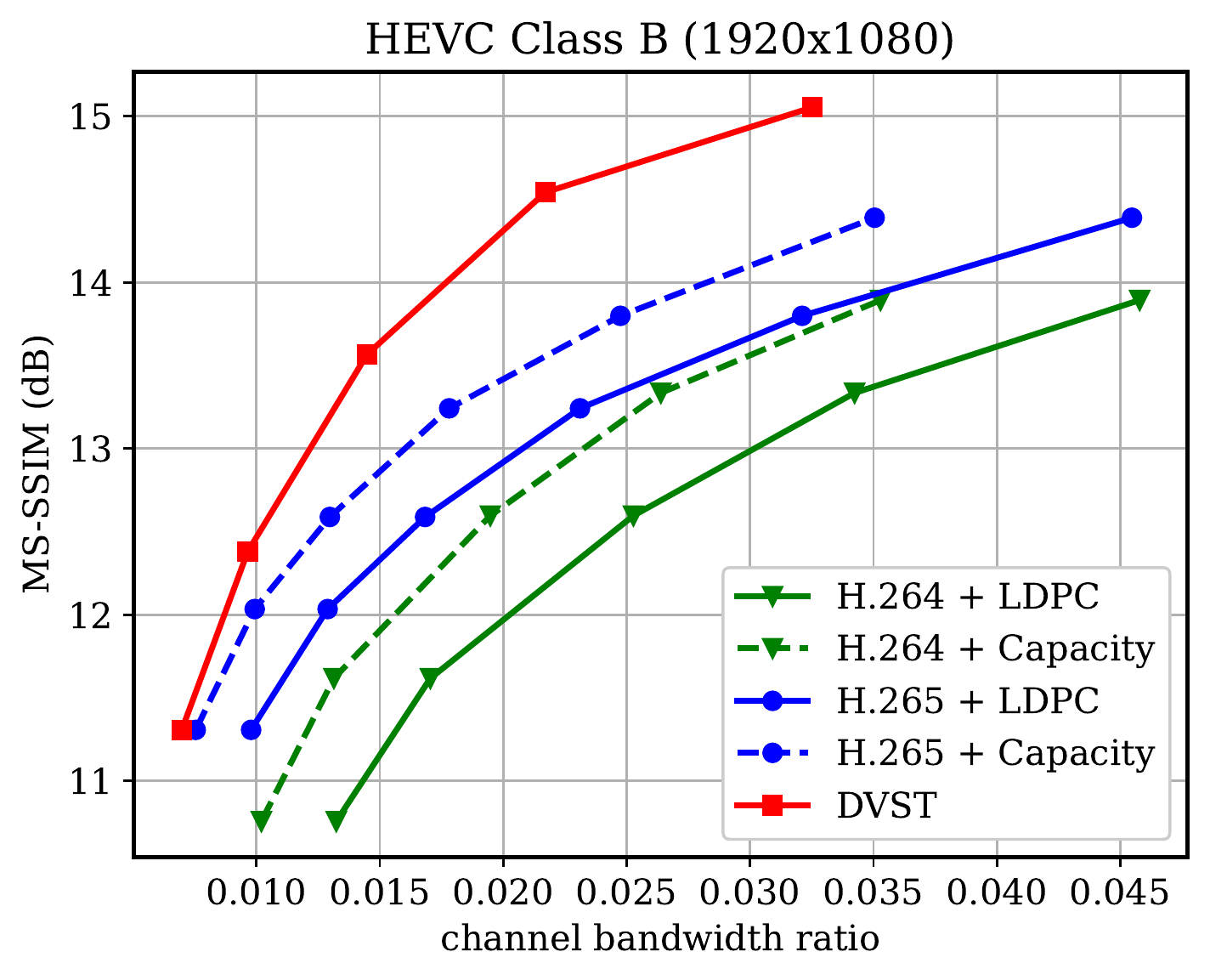}
		}
        \hspace{-.025in}
		\quad
		\subfigure[]{
			\includegraphics[scale=0.35]{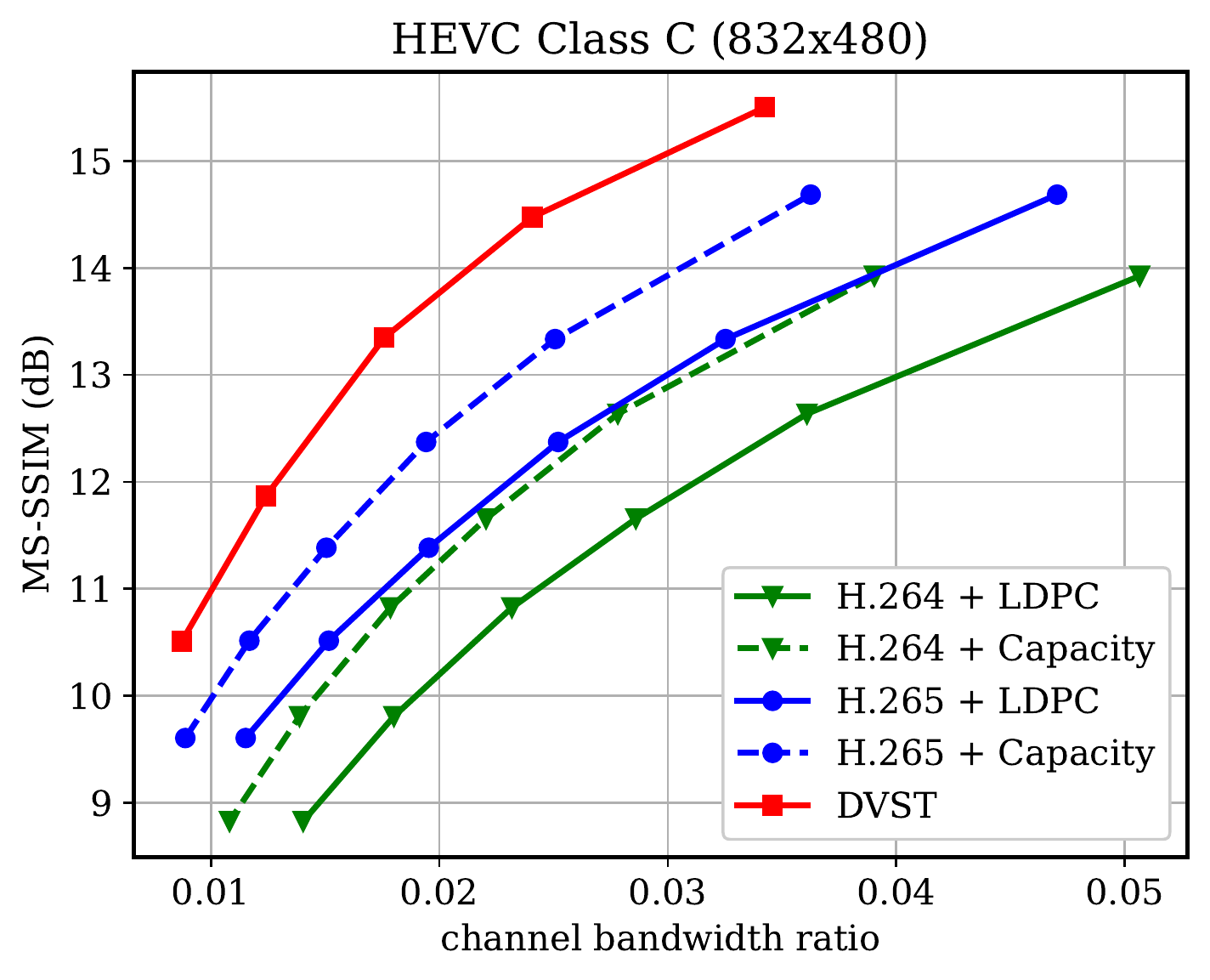}
		}
		\\
		\subfigure[]{
			\includegraphics[scale=0.35]{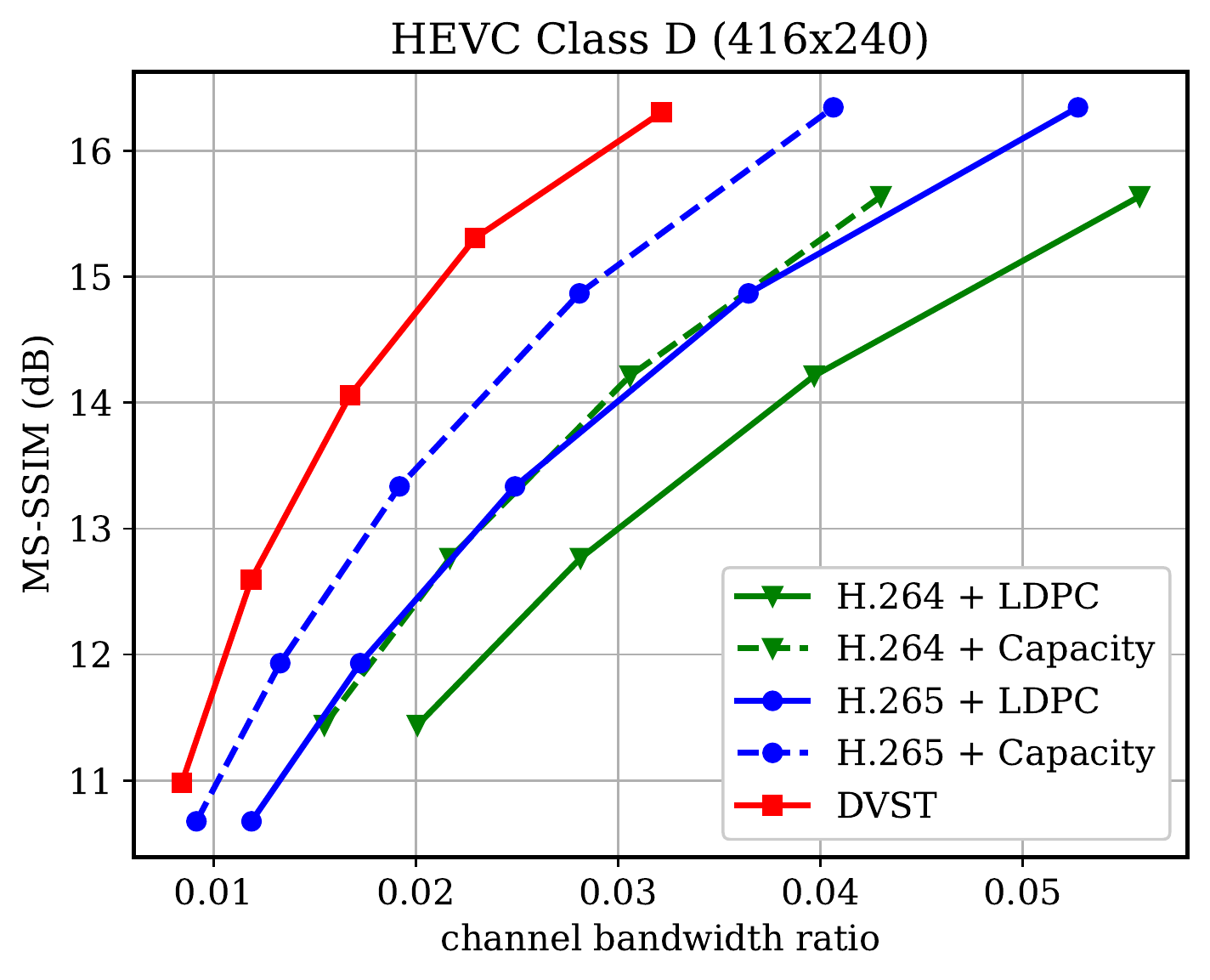}
		}
		\quad
		\subfigure[]{
			\includegraphics[scale=0.35]{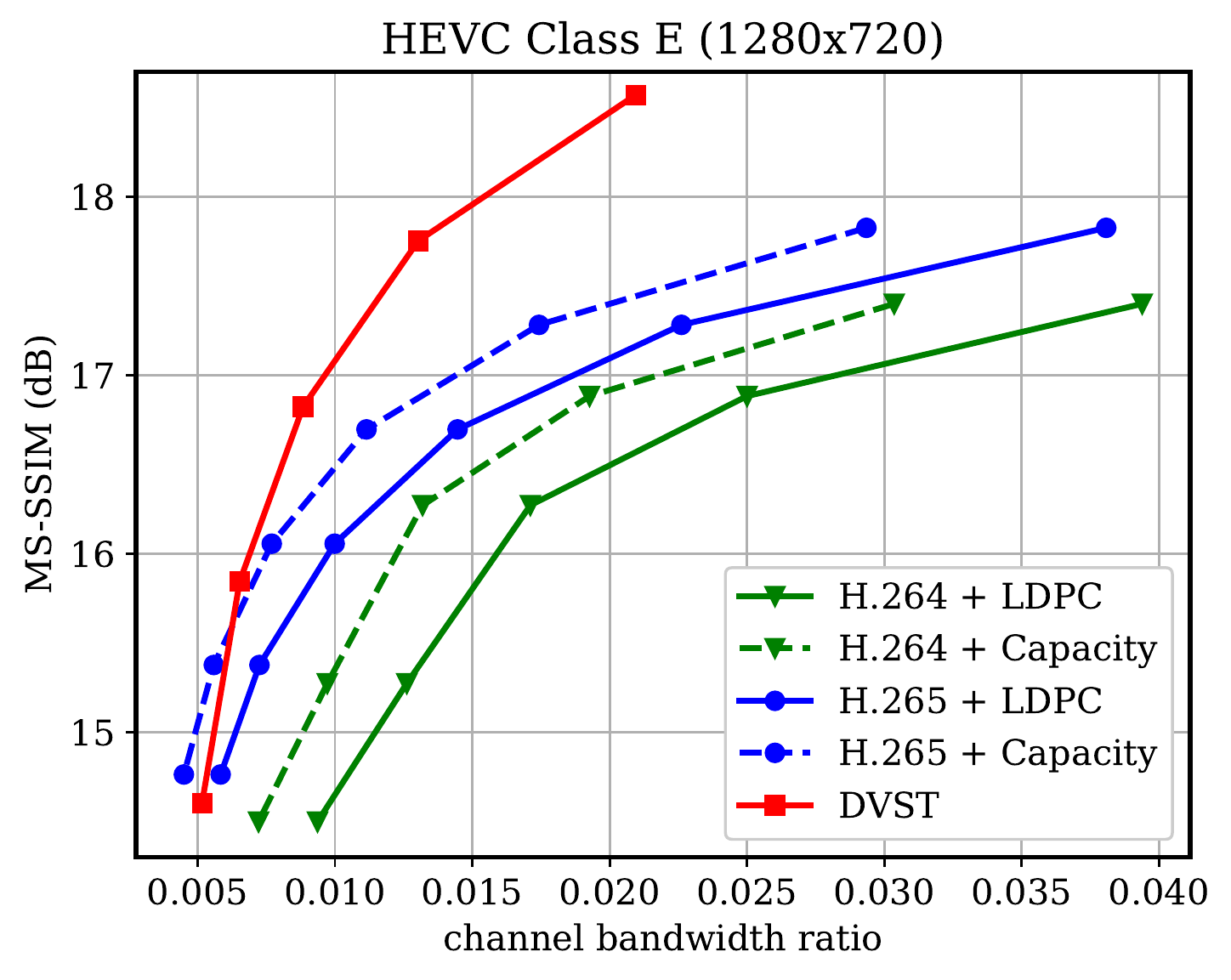}
		}
        \quad
		\subfigure[]{
			\includegraphics[scale=0.35]{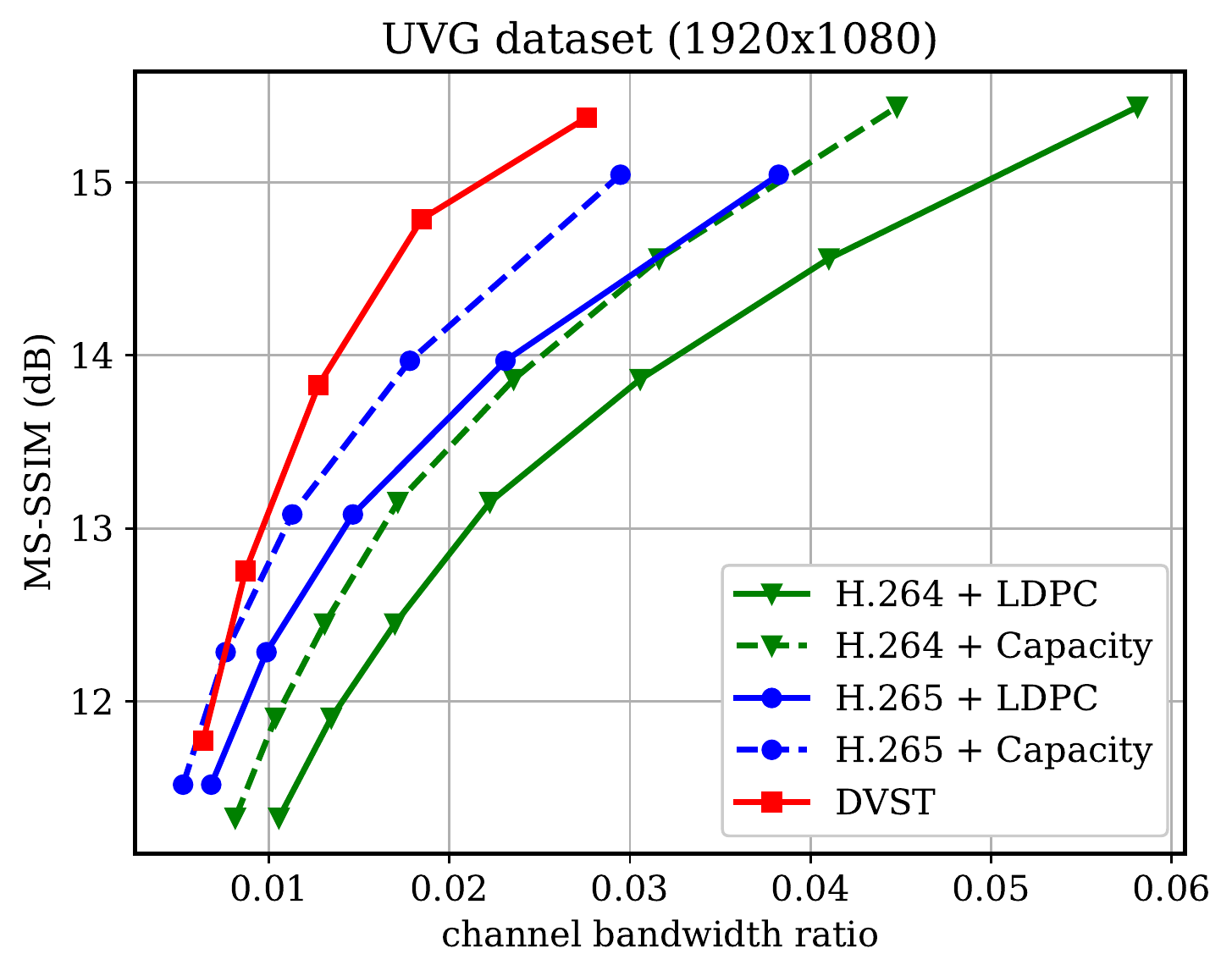}
		}
		\caption{MS-SSIM performance versus the average channel bandwidth ratio (CBR) over the AWGN channel at $ \text{SNR} = 10$dB. All MS-SSIM values are converted to dB ($-10\log_{10}(1 - m)$, where $m$ is the MS-SSIM value in the range between 0 and 1).}
		\label{Fig7}
	\end{center}
\end{figure*}

\begin{figure*}[t]
	\setlength{\abovecaptionskip}{0.cm}
	\setlength{\belowcaptionskip}{-0.cm}
	\begin{center}
		\hspace{-.0in}
		\subfigure[]{	\includegraphics[scale=0.28]{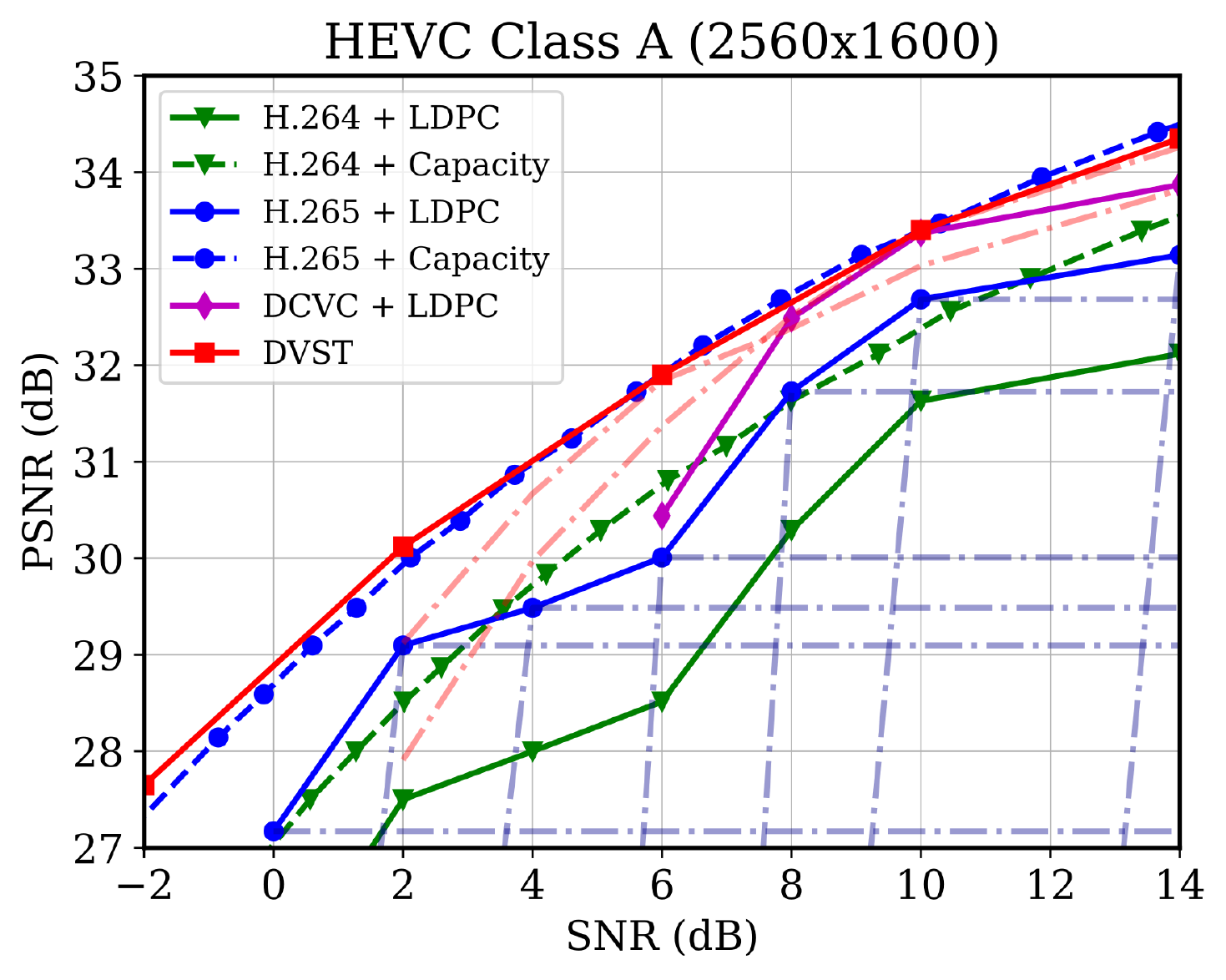}
		}
		\hspace{-.2in}
		\quad
		\subfigure[]{
			\includegraphics[scale=0.28]{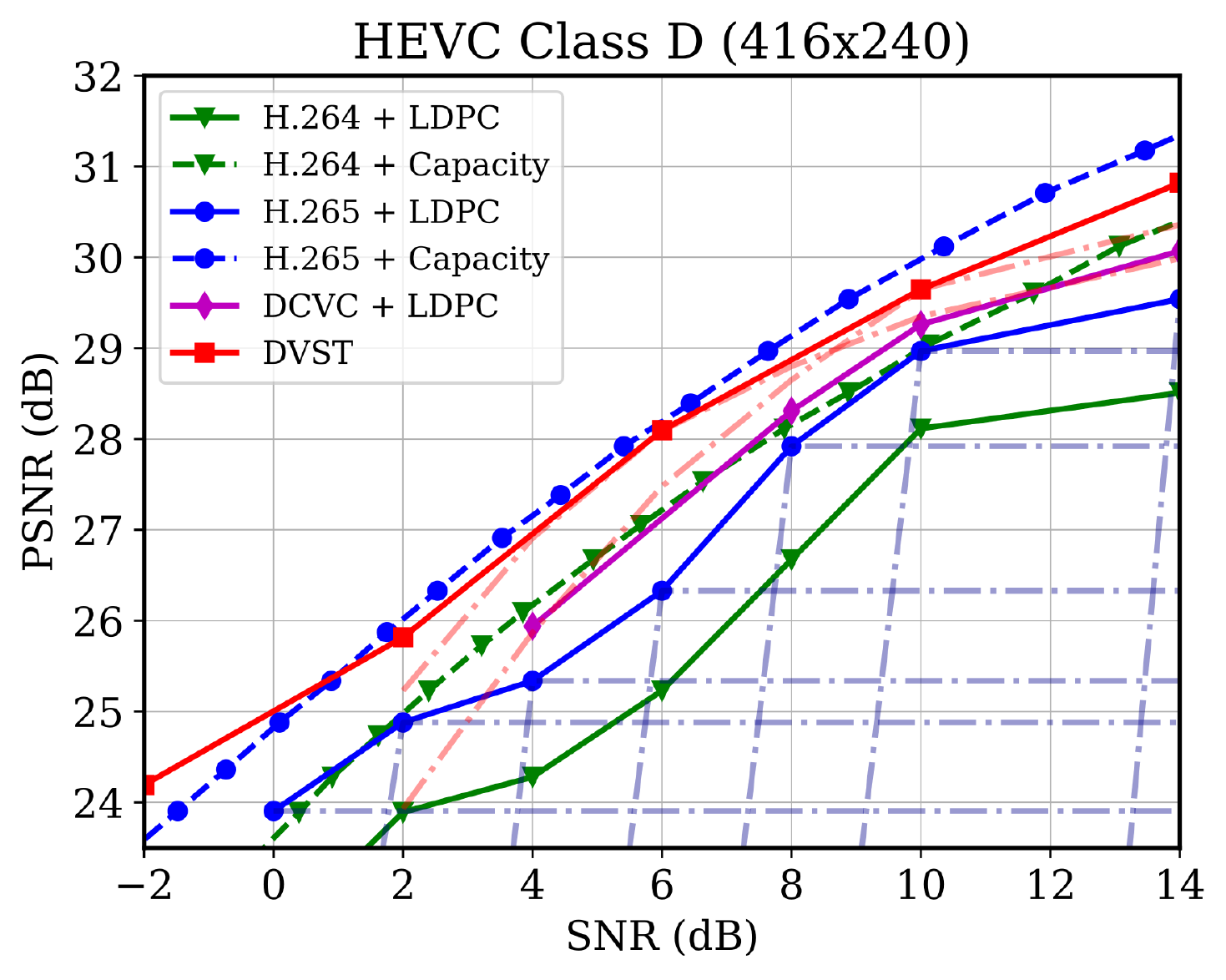}
		}
        \quad
        \hspace{-.2in}
		\subfigure[]{	\includegraphics[scale=0.28]{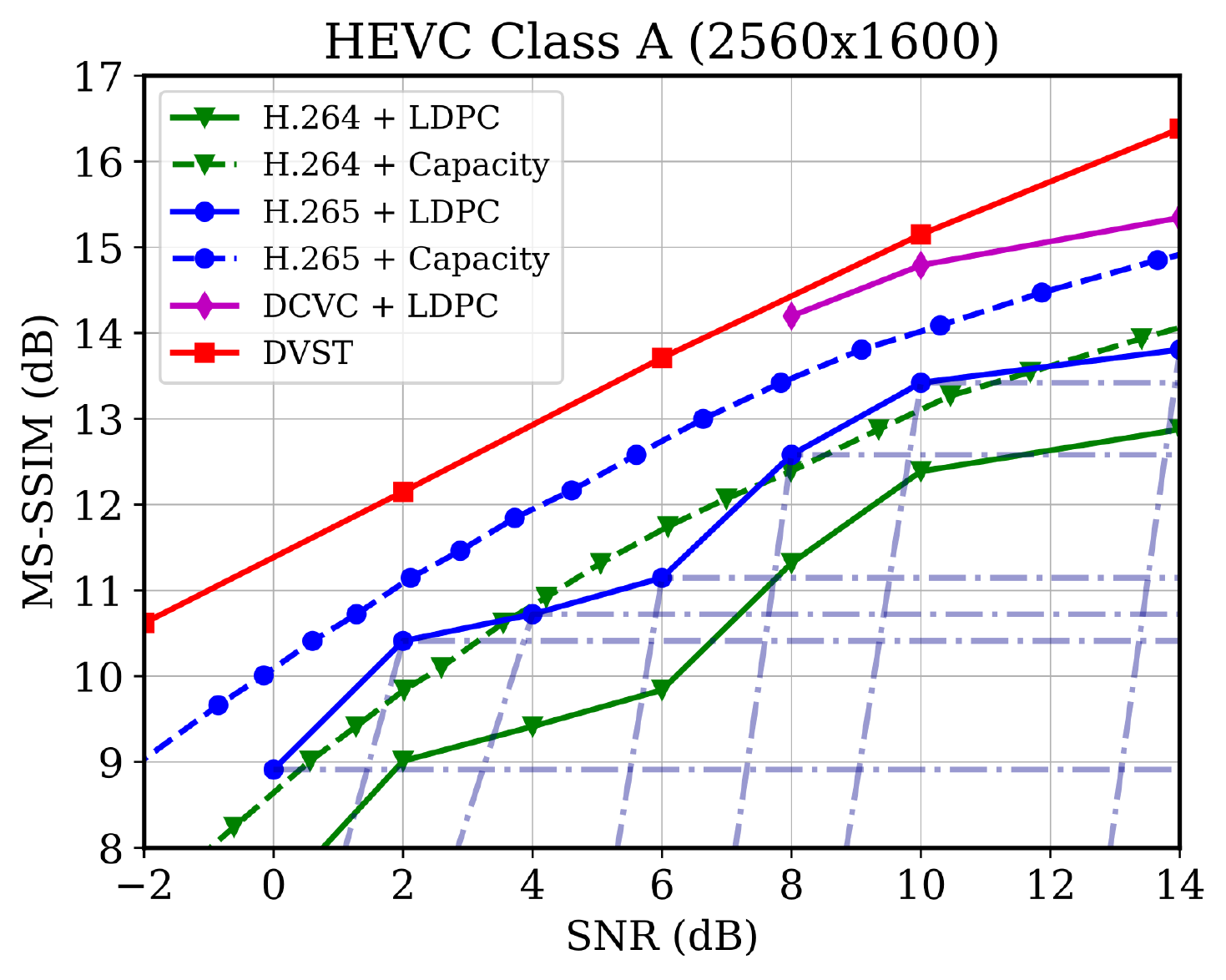}
		}
		\hspace{-.2in}
		\quad
		\subfigure[]{
			\includegraphics[scale=0.28]{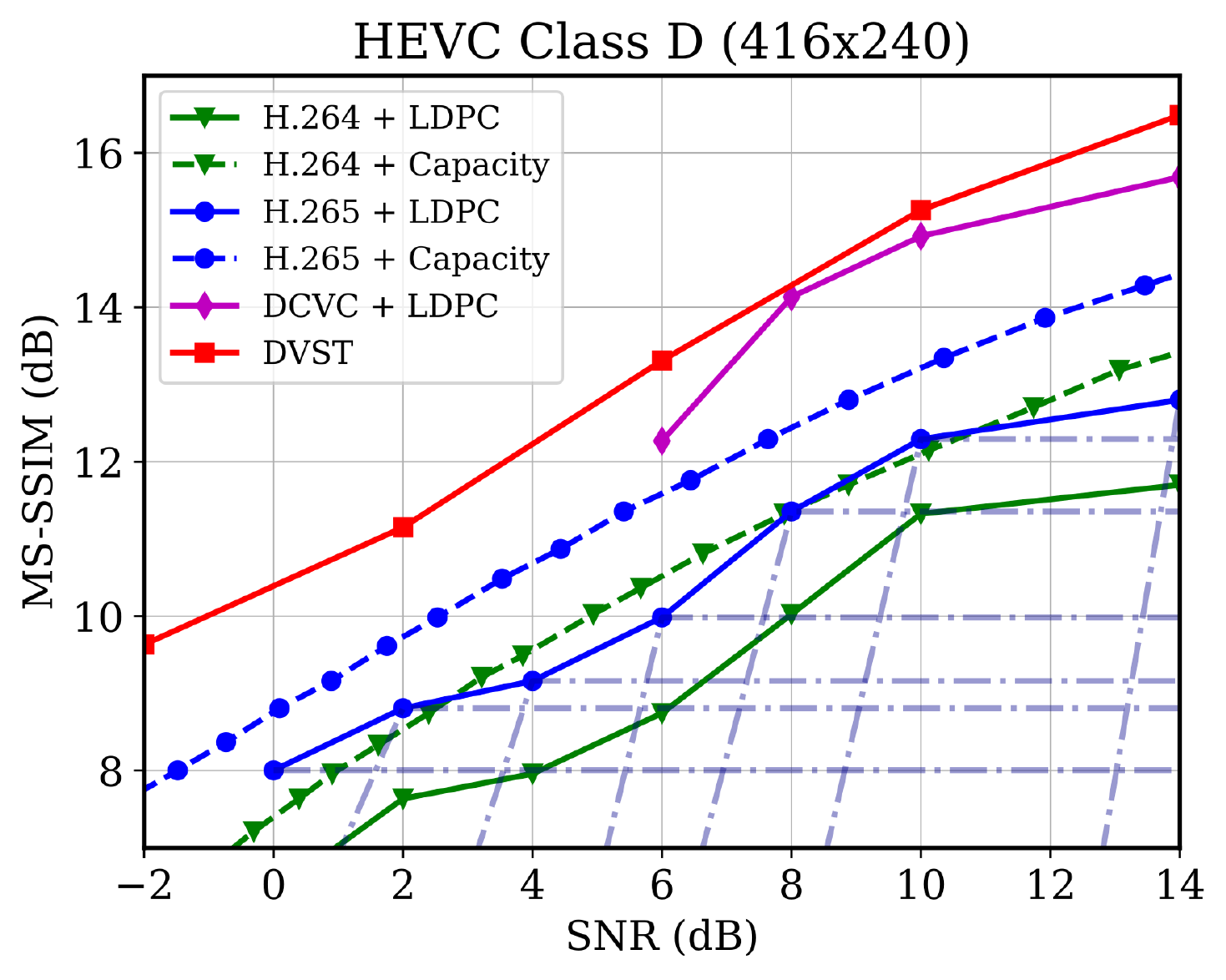}
		}
		\caption{PSNR and MS-SSIM performance versus the change of channel SNR over the AWGN channel. The CBR constraint for HEVC Class A sequence is set to $ R = 0.015$ and $ R = 0.025$ for Class D sequence.}
		\label{Fig8}
	\end{center}
\end{figure*}

Fig. \ref{Fig6} shows the RD results under the PSNR metric among various test sequences over the AWGN channel with channel $ \text{SNR} = 10$dB. For H.264 + LDPC and H.265 + LDPC, after traversing given combinations of LDPC coded modulation schemes, we exploit a $2/3$ rate $(4096, 6144)$ LDPC code with 16-ary quadrature amplitude modulation (16QAM) to ensure reliable transmission and the highest efficiency \cite{DJSCCF}. We use the Gaussian capacity formula \cite{Shannon1948A} to calculate the maximum transmission rate per channel symbol for the ideal H.264 + Capacity and H.265 + Capacity schemes. From Fig. \ref{Fig6}, we can find that on most test sequences, the proposed DVST (PSNR) scheme can outperform the practical H.264 + LDPC scheme by a large margin for all CBRs, and the performance gap increases with the CBR which indicates a better coding gain achieved by our DVST method. Furthermore, the proposed DVST shows competitive performance to the H.265 + LDPC scheme and even performs close to H.265 + Capacity in some test sequences.

As for the coding gain that shows as the RD curve slope, by using the adaptive rate allocation and contextual transmission mechanism, our DVST model shows comparable coding gain as that of H.265/H.264 series in most cases. The coding gain generally increases with the resolution of the video sequence, which demonstrates the potential of DVST on transmitting higher resolution videos over wireless channels. However, we also note that DVST performs slightly worse than H.265 + LDPC on HEVC Class C and E. A possible reason is that many video sequences, e.g., BQMall, in these two Classes consist of complex foreground or various textures, which results in difficulty to the context generation in both semantic feature space and deep JSCC codeword space. As a comparison, our DVST performs better than H.265 + LDPC on HEVC Class A and D, which are of relatively flat foreground and simple textures.

Fig. \ref{Fig7} shows the RD performance in terms of MS-SSIM perceptual metric over the AWGN channel at $\text{channel SNR} = 10$dB. Since MS-SSIM yields values between 0 (worst) and 1 (best), and most values are higher than 0.9, we converted the MS-SSIM values in dB to improve the legibility. For semantic communications, this perceptual metric aligns better with human feeling. Results indicate that the proposed DVST method can outperform classical schemes by a large margin, and it achieves a greater improvement on high-resolution images and high CBR regions. Compared to the PSNR results in Fig. \ref{Fig6}, we can find that traditional video coded transmission series are inferior to the learning-based DVST because traditional video compression is designed to be optimized for squared error with hand-selected constraints.

Fig. \ref{Fig8} provides the PSNR and MS-SSIM results versus the change of channel SNR, where the CBR constraint for HEVC Class A sequence is $ R = 0.015$, and $ R = 0.025$ for Class D sequence.
Since DVST learns an adaptive bandwidth allocation strategy depending on the video content and channel condition, it is difficult to strictly constrain the CBR to the predetermined value. In practice, based on the 10dB DVST models ($\lambda=64$ for PSNR, and $\lambda=1/32$ for MS-SSIM), we finetune $\eta_t$ to meet the CBR constraint in different SNRs.
For comparison schemes, we evaluate the performance using all possible combinations of $(4096, 8192)$, $(4096, 6144)$, and $(2048, 6144)$ LDPC codes with 4QAM, 16QAM, and 64QAM modulations. The solid blue line presents the envelope of the best performing configurations of H.265 + LDPC at each SNR.
In Fig. \ref{Fig8}, we use the solid red line to illustrate the performance of the DVST model trained with channel SNR at $-$2dB, 2dB, 6dB, 10dB, and 14dB, where the testing SNR equals to the training SNR. We also provide the performance of mismatched training and testing as the red dashed lines, where two models are trained under channel SNR 6dB and 10dB, respectively, but tested for various SNRs. We can find that the proposed DVST brings considerable performance gain. Comparing the three red lines, we observe that our DVST model also shows reasonable performance improvement with the increase of ${\text{SNR}}_{\text{test}}$ when ${\text{SNR}}_{\text{test}} > {\text {SNR}}_{\text {train}}$, and avoids catastrophic degradation when ${\text {SNR}}_{\text {test}} < {\text {SNR}}_{\text {train}}$. In contrast, traditional separation-based video coded transmission schemes show significant cliff effect which are plotted as the PSNR-SNR curves of H.265 + LDPC in the blue dashdotted lines.

As for the slope of each curve in Fig. \ref{Fig8}, our DVST shows better performance and comparable coding gain with that of H.264/H.265 series, especially on the high-resolution videos of HEVC Class A since it contains more high-frequency contents. Furthermore, we compare our DVST with the emerging neural video compression scheme DCVC \cite{DCVC} combined with LDPC codes for wireless transmission. For fair comparison, we use the same I-frame coding and GOP size and only compare the P-frame performance. Compared with DCVC + LDPC, we also achieve meaningful gain, which indicates our DVST can benefit from the good match between the learned deep JSCC and the nonlinear transform. Moreover, DVST does not rely on explicit entropy coding for compression and channel codes for error-correcting, which avoids the cliff effect and reduces the computational complexity, but DCVC + LDPC will also involve the cliff effect due to the use of entropy coding and left errors in LDPC decoding.

\begin{figure}[t]
	\setlength{\abovecaptionskip}{0.cm}
	\setlength{\belowcaptionskip}{-0.cm}
	\begin{center}
		\includegraphics[scale=0.5]{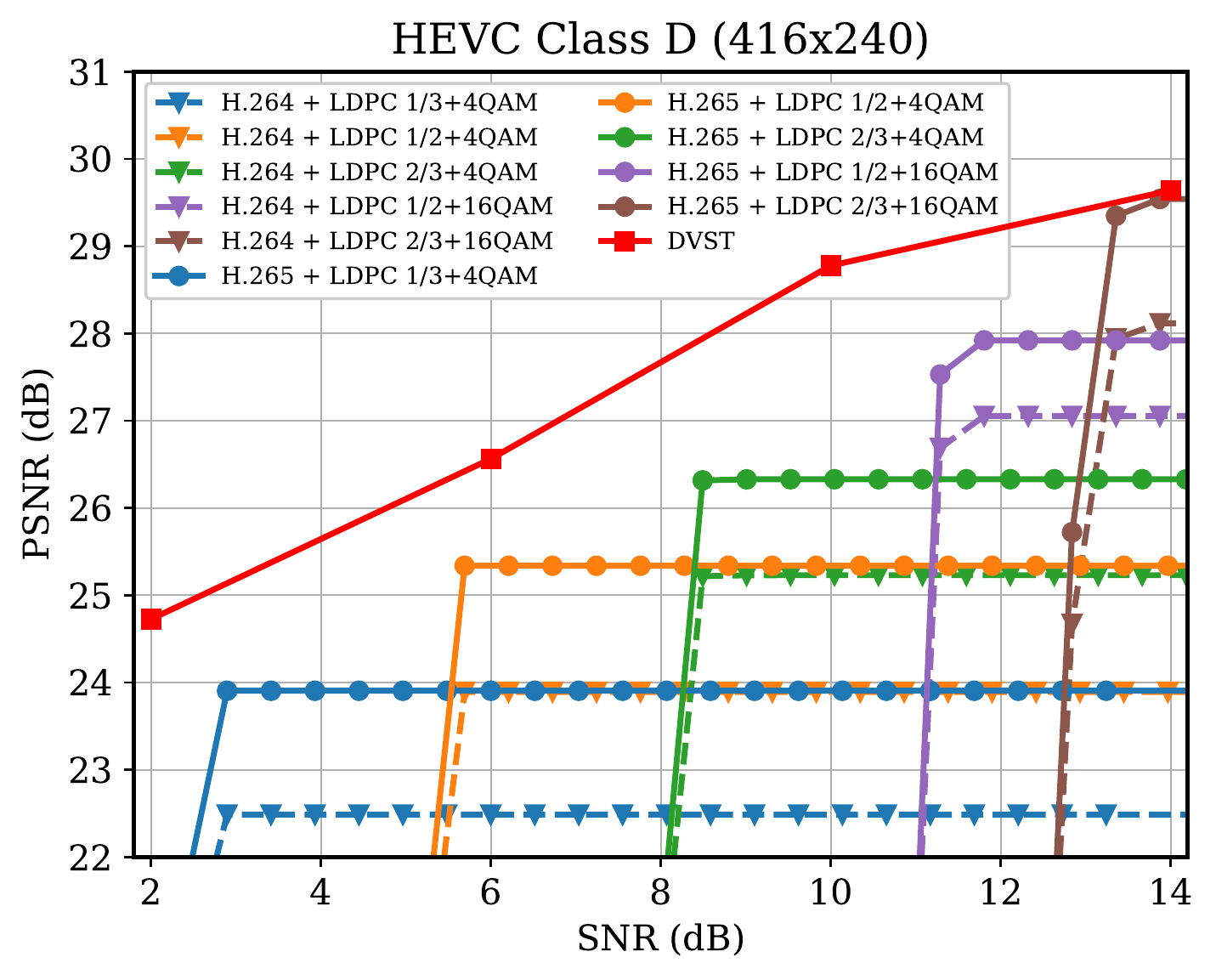}
		\caption{PSNR performance versus channel SNR over the Rayleigh fading channel. The CBR constraint for HEVC Class D sequence is set to $ R = 0.025$.}
		\label{Fig9}
	\end{center}
\end{figure}

\begin{figure*}[t]
	\setlength{\abovecaptionskip}{0.cm}
	\setlength{\belowcaptionskip}{-0.cm}
	\begin{center}
		\subfigure[]{	
			\includegraphics[scale=0.28]{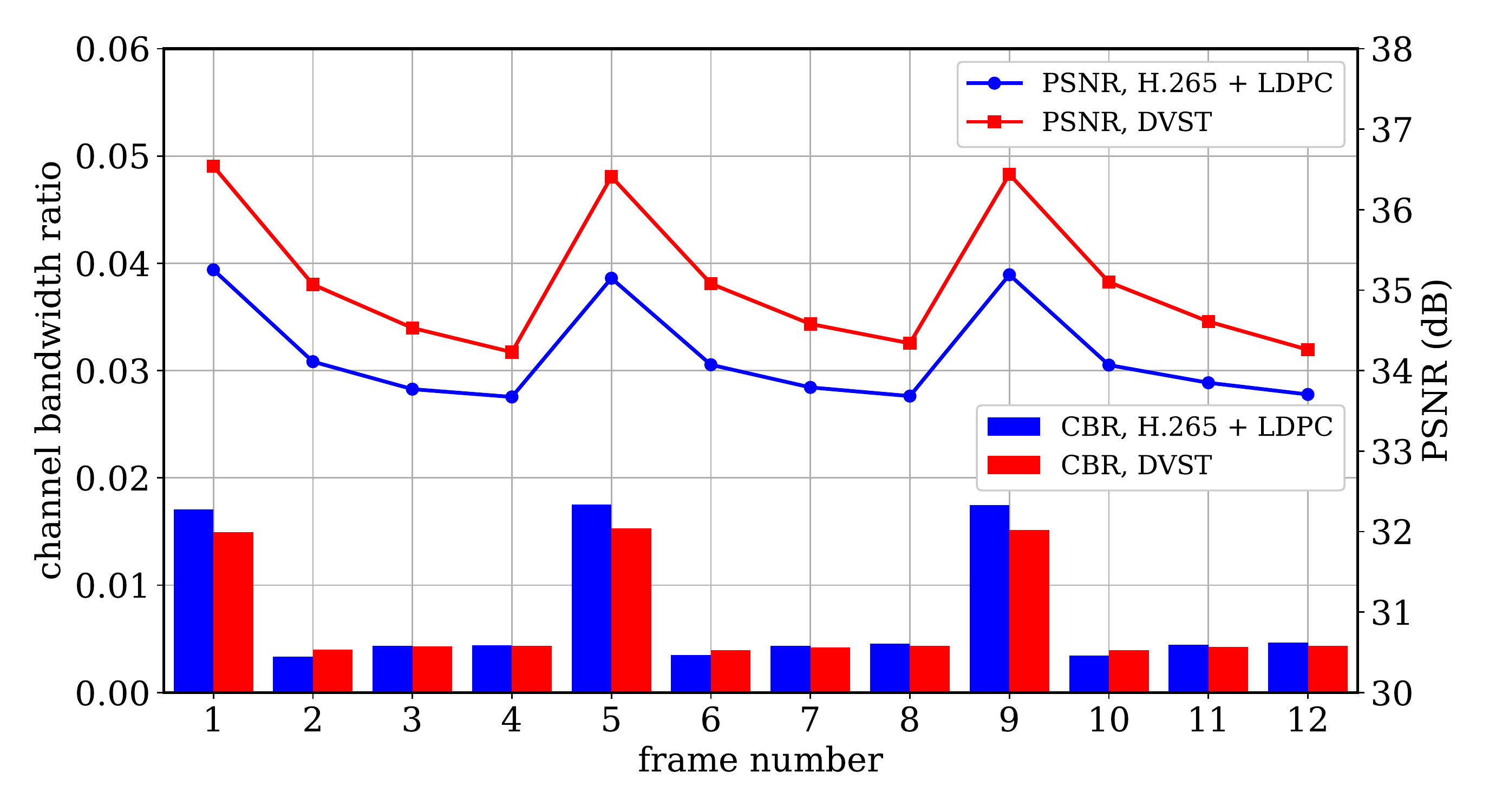}
		}
        \quad
		\subfigure[]{
			\includegraphics[scale=0.28]{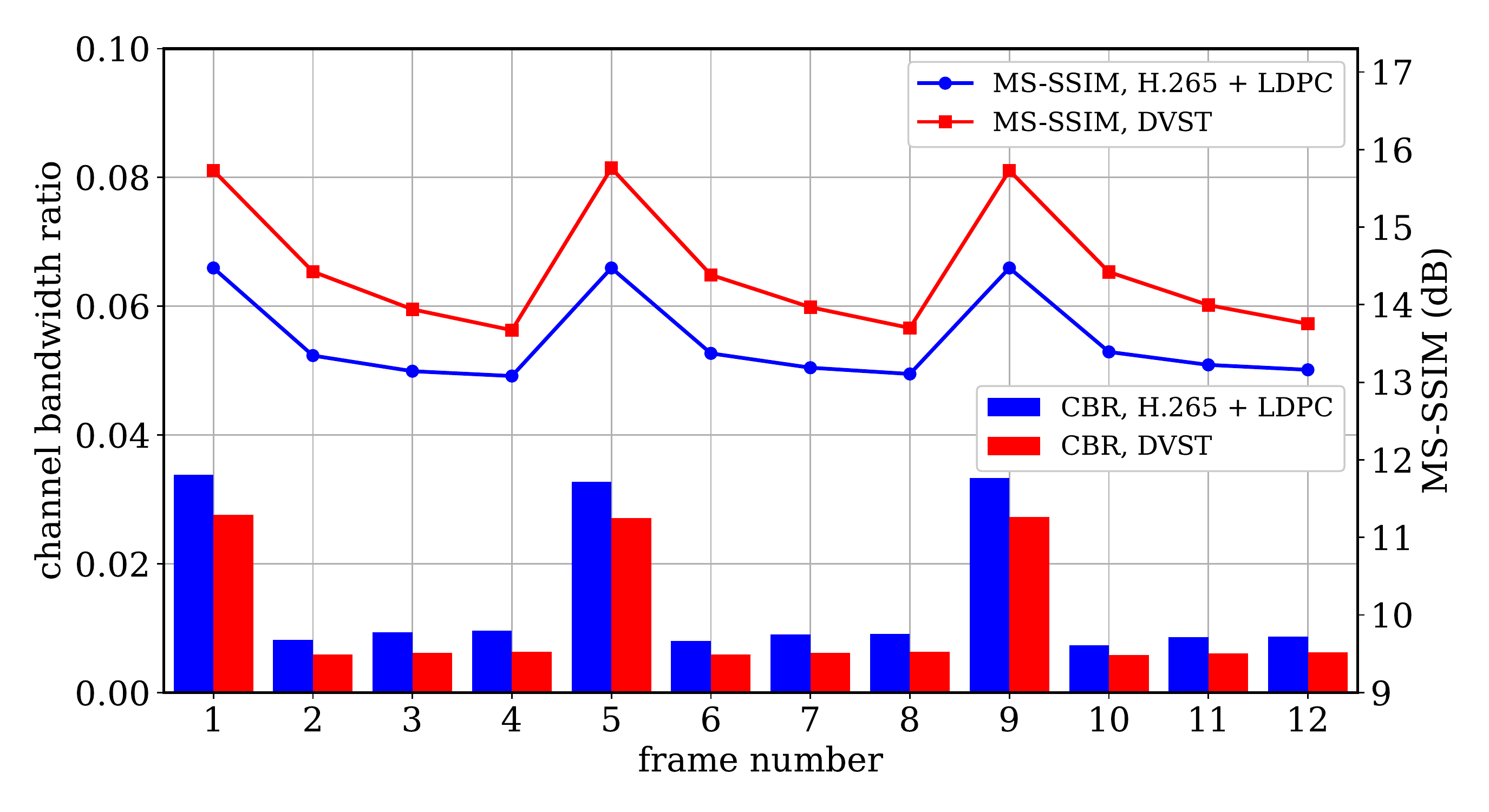}
		}
		\caption{CBR and PSNR/MS-SSIM comparison between DVST and H.265 + LDPC. We use the last 12 frames of \emph{Kimono} from HEVC Class B dataset as an example. }
		\label{Fig10}
	\end{center}
\end{figure*}

\begin{figure*}[t]
	\setlength{\abovecaptionskip}{0.cm}
	\setlength{\belowcaptionskip}{-0.cm}
	\begin{center}
		\includegraphics[scale=0.38]{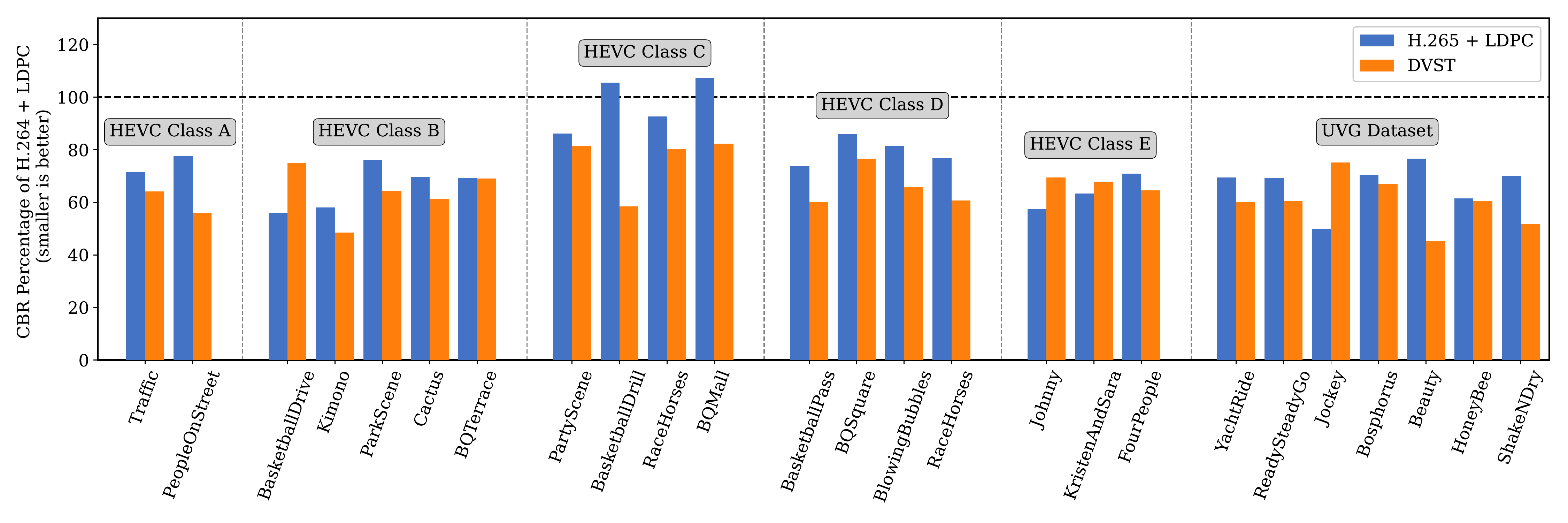}
		\caption{CBR savings for each video in all test sequences. We use the widely used Bjontegaard Delta (BD) rate reduction \cite{BDrate} algorithm to estimate the bandwidth savings. The rate saving value represents the percentage of channel bandwidth cost relative to H.264 + LDPC at the same PSNR (smaller is better), and this experiment is conducted at the AWGN channel with $\text{SNR}=10$dB. As the anchor, H.264 + LDPC scheme is always $100\%$ as plotted in the dashed horizontal line. In most cases ($21/25$), the proposed DVST can save more bandwidth compared to both H.264 + LDPC and H.265 + LDPC.}
		\label{Fig11}
	\end{center}
\end{figure*}

\makeatletter
\renewcommand{\@thesubfigure}{\hskip\subfiglabelskip}
\makeatother

\begin{figure*}

 \begin{subtable}
  \centering
  \small
  \begin{tabular}{m{0.20\textwidth}m{0.17\textwidth}<{\centering}m{0.16\textwidth}<{\centering}m{0.16\textwidth}<{\centering}m{0.16\textwidth}<{\centering}<{\centering}m{0.16\textwidth}<{\centering}}
   & Original & H.264 + LDPC &  H.265 + LDPC &  DVST
  \end{tabular}
 \end{subtable}
 \begin{center}

 \hspace{-.05in}
 \subfigure[] {\includegraphics[width=0.18\textwidth]{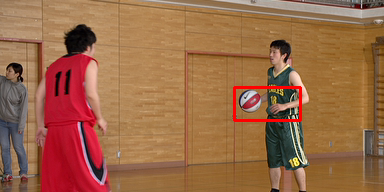}}
 \hspace{-.10in}
 \quad
 \subfigure[$R_{\text{GOP}}$ / PSNR (dB)] {\includegraphics[width=0.18\textwidth]{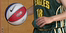}}
 \hspace{-.20in}
 \quad
 \subfigure[0.024 (\textit{0\%}) / 31.35] {\includegraphics[width=0.18\textwidth]{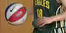}}
 \hspace{-.20in}
 \quad
 \subfigure[0.022 (\textcolor{blue}{\textit{--8.3\%}}) / 31.93]{\includegraphics[width=0.18\textwidth]{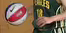}}
 \hspace{-.20in}
 \quad
 \subfigure[0.015 (\textcolor{blue}{\textit{--37.5\%}}) / 32.12] {\includegraphics[width=0.18\textwidth]{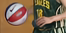}}

 \hspace{-.05in}
 \subfigure[] {\includegraphics[width=0.18\textwidth]{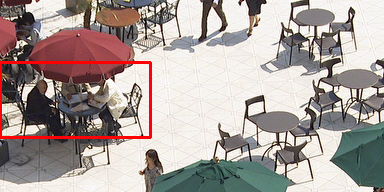}}
 \hspace{-.10in}
 \quad
 \subfigure[$R_{\text{GOP}}$ / MS-SSIM (dB)] {\includegraphics[width=0.18\textwidth]{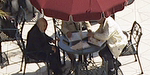}}
 \hspace{-.20in}
 \quad
 \subfigure[0.018 (\textit{0\%}) / 12.79] {\includegraphics[width=0.18\textwidth]{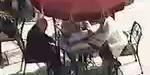}}
 \hspace{-.20in}
 \quad
 \subfigure[0.017 (\textcolor{blue}{\textit{--5.5\%}}) / 13.02]{\includegraphics[width=0.18\textwidth]{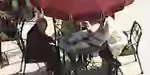}}
 \hspace{-.20in}
 \quad
 \subfigure[0.016 (\textcolor{blue}{\textit{--11.1\%}}) / 15.91] {\includegraphics[width=0.18\textwidth]{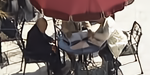}}

 \caption{Examples of visual comparison. The first column shows the original frame. The second column shows the cropped patch in original frame. The third to the fifth column show the reconstructed frames by using different transmission schemes over the AWGN channel at $\text{SNR} = 10$dB. Note that, the fifth column presents the cropped patch generated by DVST (PSNR) and DVST (MS-SSIM), respectively. $R_{\text{GOP}}$ denotes the average CBR of the current GOP. The blue number indicates the percentage of bandwidth cost saving compared to the baseline ``H.264 + LDPC'' scheme.}
 \label{Fig12}
\end{center}

\end{figure*}

\begin{figure*}
	
	\begin{subtable}
		\centering
		\small
		\begin{tabular}{m{0.20\textwidth}m{0.175\textwidth}<{\centering}m{0.157\textwidth}<{\centering}m{0.159\textwidth}<{\centering}m{0.159\textwidth}<{\centering}<{\centering}m{0.16\textwidth}<{\centering}}
			&$\text{SNR}_\text{test} = 2\text{dB}$ & $\text{SNR}_\text{test} = 6\text{dB}$ & $\text{SNR}_\text{test} = 10\text{dB}$ & $\text{SNR}_\text{test} = 14\text{dB}$  \\
		\end{tabular}
	\end{subtable}
	\begin{center}
		
		\hspace{-.05in}
		\subfigure[] {\includegraphics[width=0.18\textwidth]{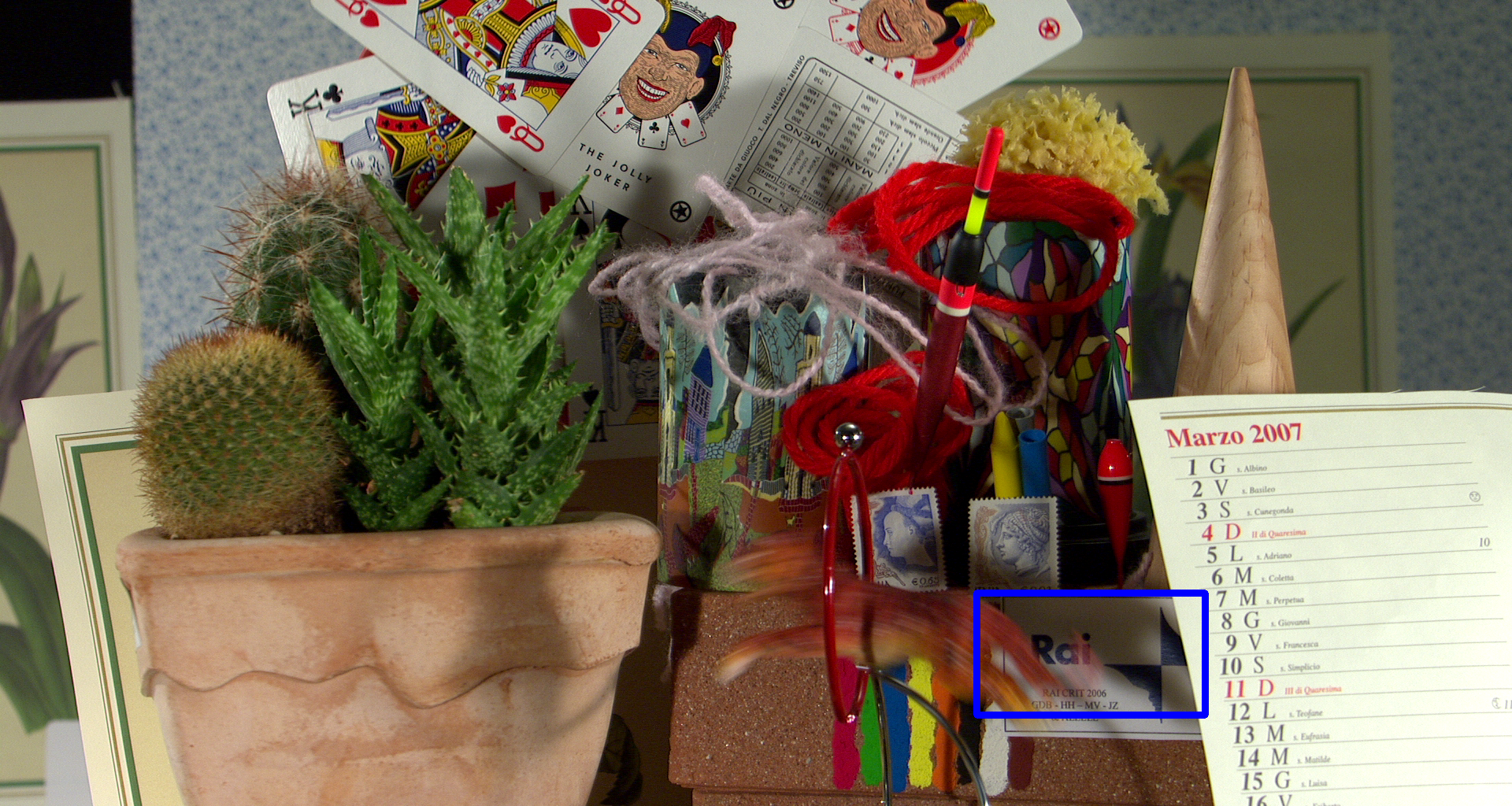}}
		\hspace{-.10in}
		\quad
		\subfigure[DVST ($\text{SNR}_\text{train} = 2\text{dB}$)] {\includegraphics[width=0.18\textwidth]{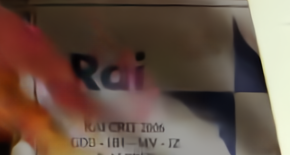}}
		\hspace{-.20in}
		\quad
		\subfigure[DVST ($\text{SNR}_\text{train} = 6\text{dB}$)] {\includegraphics[width=0.18\textwidth]{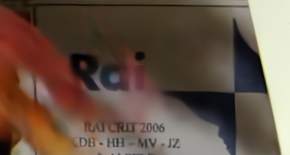}}
		\hspace{-.20in}
		\quad
		\subfigure[DVST ($\text{SNR}_\text{train} = 10\text{dB}$)]{\includegraphics[width=0.18\textwidth]{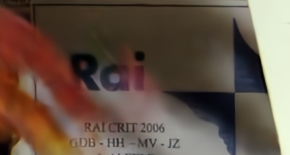}}
		\hspace{-.20in}
		\quad
		\subfigure[DVST ($\text{SNR}_\text{train} = 14\text{dB}$)] {\includegraphics[width=0.18\textwidth]{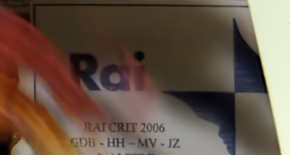}}
		
		\hspace{-.05in}
		\subfigure[Original] {\includegraphics[width=0.18\textwidth]{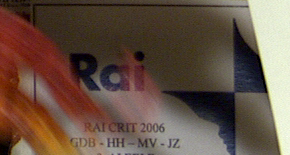}}
		\hspace{-.10in}
		\quad
		\subfigure[H.265+1/2 LDPC+4QAM] {\includegraphics[width=0.18\textwidth]{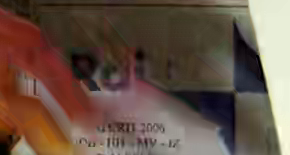}}
		\hspace{-.20in}
		\quad
		\subfigure[H.265+2/3 LDPC+4QAM] {\includegraphics[width=0.18\textwidth]{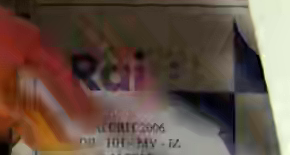}}
		\hspace{-.20in}
		\quad
		\subfigure[H.265+2/3 LDPC+16QAM]{\includegraphics[width=0.18\textwidth]{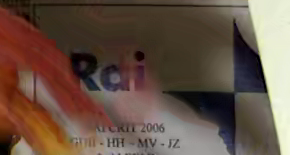}}
		\hspace{-.20in}
		\quad
		\subfigure[H.265+1/2 LDPC+64QAM] {\includegraphics[width=0.18\textwidth]{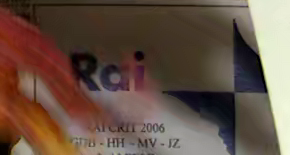}}
		
		\caption{Examples of visual comparison. The first column shows the original frame and its cropped patch. The second to the fifth column shows the frames reconstructed by DVST (PSNR) or H.265 + LDPC over the AWGN channel at various channel SNRs, respectively. For the DVST (PSNR), we consider the condition when ${\text{SNR}}_{\text{test}} = {\text {SNR}}_{\text {train}}$. For the H.265 + LDPC scheme, the rate of LDPC and the order of QAM modulation is marked below the reconstructed frame. The average CBR of GOP is limited to 0.012.}
		\label{Fig13}
	\end{center}
	
\end{figure*}

Next, we show the PSNR performance under the Rayleigh fading channel with CBR constraint $R=0.025$ in Fig. \ref{Fig9}. In this case, we assume the Rayleigh fading channel gain vector $\mathbf{h}_t \sim \mathcal{CN}(\mathbf{0}, \mathbf{I}_{k_t})$, and it is known at the receiver with ideal channel estimation. Thus, the receiver first conducts channel equalization, such that the received signal can be equivalently written as $\mathbf{\hat{s}}_t = \mathbf{s}_t + {\mathbf{n}}_t / {\mathbf{h}}_t$, and then feed $\mathbf{\hat{s}}_t$ into DVST decoder. In practice, our DVST models of the Rayleigh fading channel are finetuned from baseline models learned under the AWGN channel with the same SNR. Apparently, classical separation schemes (H.264/H.265 + LDPC + QAM) are still inferior to our DVST, especially in the low SNR region.

Fig. \ref{Fig10} shows a detailed comparison between DVST and H.265 + LDPC over the performance of a group of consecutive frames. It can be observed that the reconstruction quality of both schemes degrades with the increase of P-frame number within one GOP. In comparison, our DVST can spend fewer channel bandwidth costs while achieving much better reconstruction quality.

Furthermore, as shown in Fig. \ref{Fig11}, we compute the BD rate reduction \cite{BDrate} relative to H.264 + LDPC for each video under AWGN channel at $\text {SNR = 10}$dB. Compared with H.264 + LDPC, the channel bandwidth cost of DVST is only 40\% to 80\% under the same reconstruction quality in terms of PSNR, which means a 60\% to 20\% the channel bandwidth can be saved. Compared with the results of H.265 + LDPC, DVST can still save more bandwidth in most videos ($21/25$).

Fig. \ref{Fig12} and Fig. \ref{Fig13} provide illustrative examples to demonstrate the performance of DVST intuitively. Specifically, as shown in Fig. \ref{Fig12}, we visualize the specific reconstructed frames of Fig. \ref{Fig6} and Fig. \ref{Fig7} in the two rows, respectively. From the two groups of examples, we can observe that our DVST model generates high fidelity reconstructions with lower CBR costs.
Fig. \ref{Fig13} presents the reconstruction results versus the change of channel SNR under a limited CBR budget. It can be seen that the results of H.265 + LDPC + QAM scheme have artifacts and block effects in the low SNR region, while DVST generates a clear text.

\begin{figure}[t]
	\setlength{\abovecaptionskip}{0.cm}
	\setlength{\belowcaptionskip}{-0.cm}
	\centering{\includegraphics[scale=0.5]{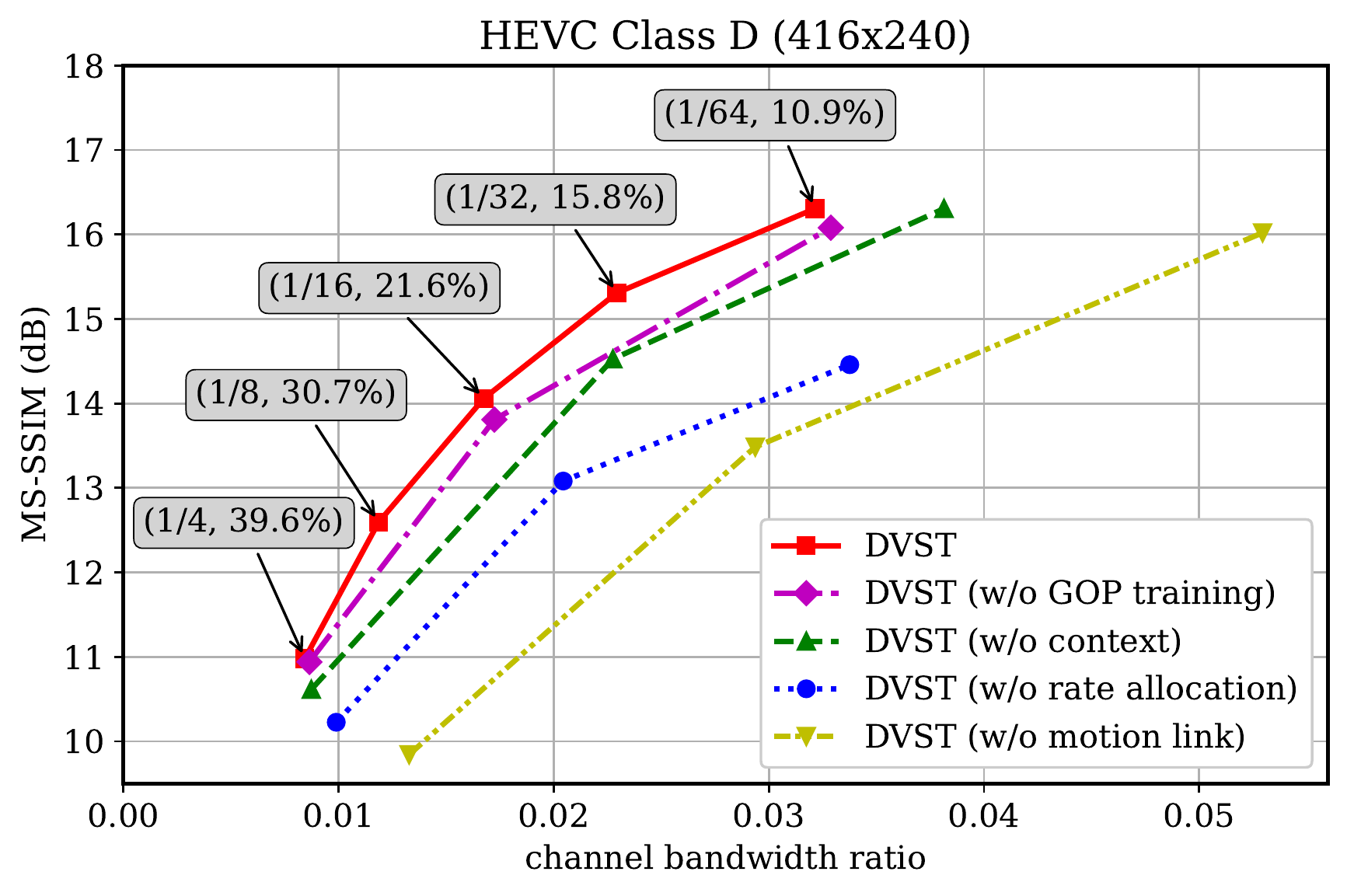}}
	\caption{Ablation study results in terms of MS-SSIM over the AWGN channel at SNR = $10$dB. The hyperparameter $\lambda$ and the CBR percentage of motion link are also marked on the DVST curve.}
	\vspace{0em}
	\label{Fig14}
\end{figure}

\begin{figure}[t]
	\setlength{\abovecaptionskip}{0.cm}
	\setlength{\belowcaptionskip}{-0.cm}
	\begin{center}
		\subfigure[(a)] {	
			\includegraphics[scale=0.34]{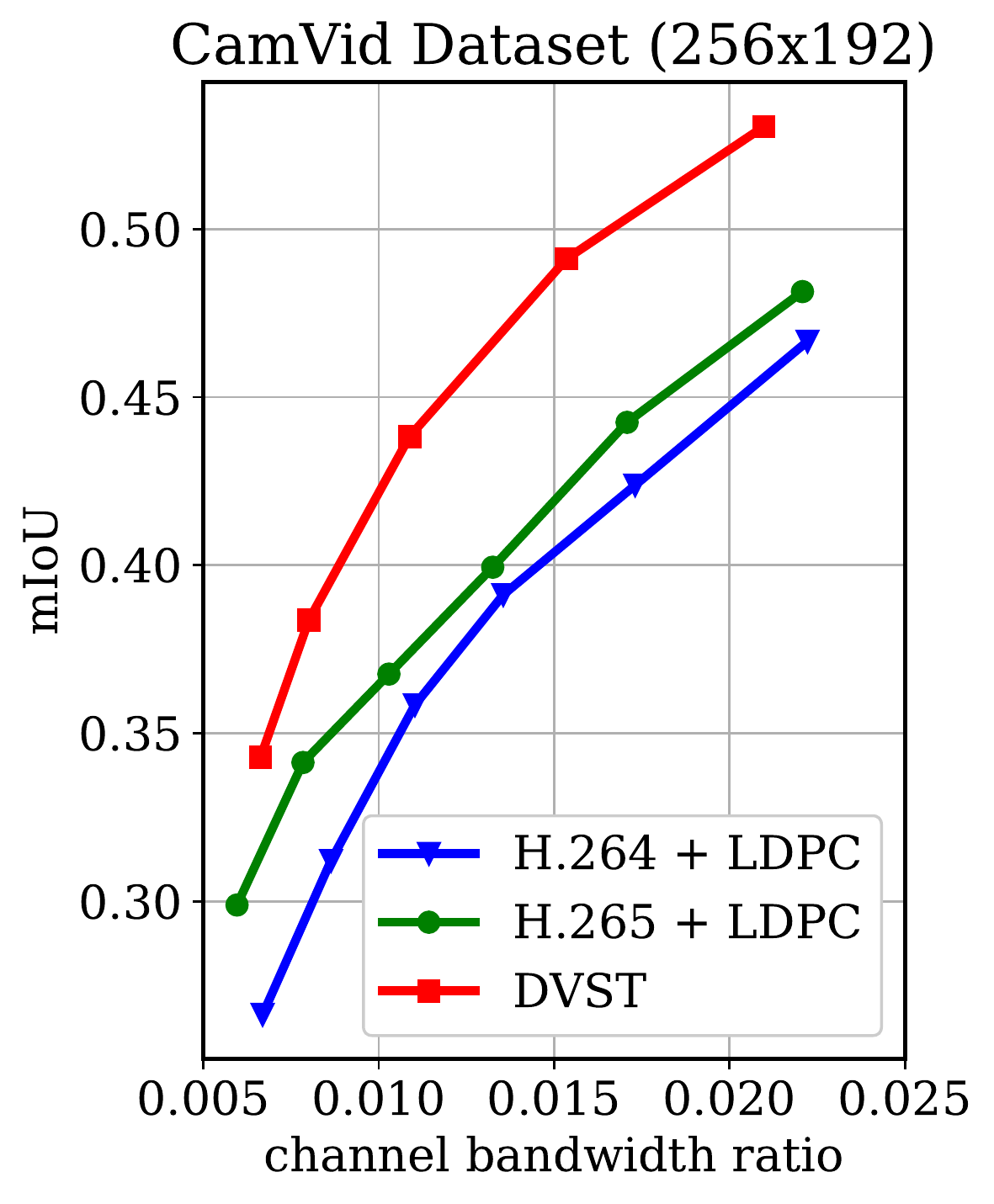}
		}
		\subfigure[(b)] {
			\includegraphics[scale=0.34]{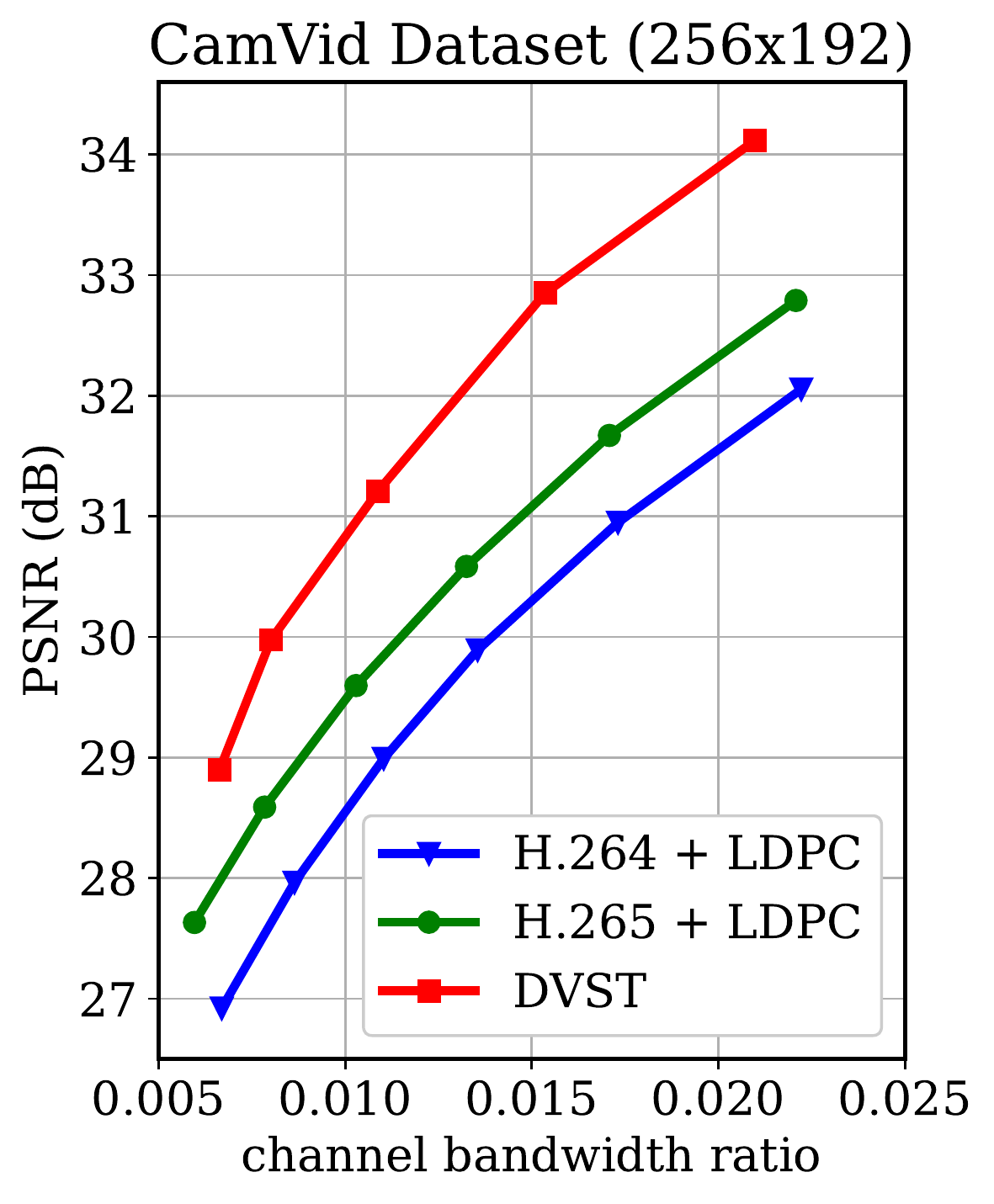}
		}
		\caption{Reconstruction and segmentation performance versus the average CBR over the AWGN channel at $ \text{SNR} = 10$dB.}
		\label{Fig15}
	\end{center}
\end{figure}

\begin{figure*}[t]
	
	\begin{subtable}
		\centering
		\small
		\begin{tabular}{m{0.20\textwidth}m{0.17\textwidth}<{\centering}m{0.16\textwidth}<{\centering}m{0.16\textwidth}<{\centering}m{0.16\textwidth}<{\centering}<{\centering}m{0.16\textwidth}<{\centering}}
		 & Ground Truth & H.264 + LDPC &  H.265 + LDPC &  DVST  \\
		\end{tabular}
	\end{subtable}
	\begin{center}
		
		\hspace{-.05in}
		\subfigure{\includegraphics[width=0.18\textwidth]{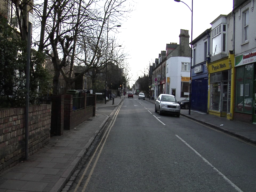}}
		\hspace{-.10in}
		\quad
		\subfigure{\includegraphics[width=0.18\textwidth]{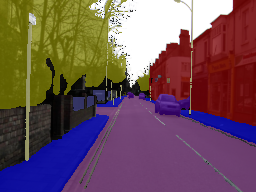}}
		\hspace{-.20in}
		\quad
		\subfigure{\includegraphics[width=0.18\textwidth]{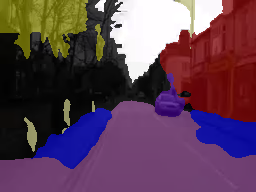}}
		\hspace{-.20in}
		\quad
		\subfigure{\includegraphics[width=0.18\textwidth]{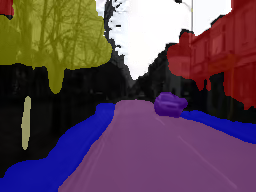}}
		\hspace{-.20in}
		\quad
		\subfigure{\includegraphics[width=0.18\textwidth]{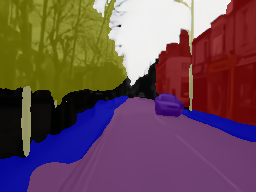}}
		
		\caption{Examples of semantic segmentation results. The first two columns present the original frame and its label. The third to the fifth column shows the segmentation results from reconstructions of H.264 + LDPC or H.265 + LDPC or DVST (PSNR), respectively. The experiment is carried out over the AWGN channel with $\text {SNR} = 10$dB and the average CBR of GOP is constrained to 0.009.}
		\label{Fig16}
	\end{center}
	
\end{figure*}

As for the ablation study, we verify the gains brought by proposed algorithms in Fig. \ref{Fig14}, including the contextual model enhancement, rate-adaptive transmission, and GOP integrated training strategy. The bandwidth cost trade-offs between primary and motion links are also provided. When the $\lambda$ becomes smaller, the DVST model tends to allocate more channel bandwidth to reduce the distortion in the primary link while the percentage of channel bandwidth in the motion link decreases. This result means it is more efficient to focus on the primary link in the high CBR region since this link can preserve more details. We verify the effectiveness of the GOP training strategy, which considers the influence of temporal correlations among adjacent frames. Comparing the results of DVST with that of DVST (without (w/o) GOP training), about 0.2dB reconstruction quality improvement can be seen. To verify the advantages of contextual video transmission, we remove the context generation network of DVST and adopt the traditional residual coding structure \cite{DVC} as an alternative, which exploits MV to warp the reference frame and transmit the residual between the wrapped frame and the input frame. This contextual model enhancement method in DVST improves the whole performance compared to traditional residual coding structure. Finally, we invalidate the rate adaption module in $f_e$, $f_e^{\rm{mv}}$, $f_d$, and $f_d^{\rm{mv}}$ by using a constant channel bandwidth cost for each patch embedding and deleting the additional rate tokens. Since the network can no longer learn a bandwidth cost trade-off between primary and motion links, the CBR proportion of the motion link is predetermined to $1/3$. Given the channel bandwidth cost for each patch embedding, the total CBR is then fixed, so the loss function of DVST (w/o rate allocation) only has the distortion term. As shown in Fig. \ref{Fig14}, our DVST overpasses the DVST (w/o rate allocation) by a large margin, especially in the high CBR region, verifying the coding gain brought by the proposed rate-adaptive transmission mechanism. Furthermore, we provide the ablation study about the motion link. The DVST (w/o motion link) extracts the context semantics solely from previous reconstructions. In this case, the whole system architecture can be vastly simplified. However, without the guidance of optical flow, it will be difficult for the context generation module to extract valuable information to exploit the spatio-temporal dependencies, resulting in significant performance degradation.

To compare the computational complexity of different video transmission systems, we measured the average encoding time of DVST on a Linux server with an Intel Xeon Gold 6226R CPU and a RTX 3090 GPU. Following the complexity analysis from \cite{DVC,DCVC}, we transmit five videos in the 1080P HEVC Class B dataset and measure the encoding speed. As a result, our DVST model spends 280ms to encode a single P-frame as channel-input symbols, which is one times faster than DCVC + LDPC, mainly due to the savings in arithmetic coding time. It is worth mentioning that the coding speed can be further improved by employing the latest deep model acceleration techniques, which is beyond the scope of this paper. As a comparison, the scheme of H.265 + LDPC runs at the speed from 1.5fps (frames per second) to 25fps with different coding settings (the trade-off between the coding efficiency and encoding speed), and the encoding speed of H.264 + LDPC is 8fps to 150fps. Note that, both H.264 and H.265 are implemented using commercial softwares with highly parallel framework and advanced assembly optimization techniques, while their official reference software runs hundreds of times slower \cite{lu2020end}.

\subsection{Downstream Machine Vision Task Results}

To potentially support future machine-type communications \cite{duan2020video}, we further optimize DVST for driving the downstream machine vision tasks while preserving the advantages of signal level reconstruction. Specifically, an analytics model is concatenated after the transmission framework to complete high-level semantics-related tasks based on the reconstructed frames. The transmission framework can be either a JSCC scheme like DVST or classical separated system like H.265 + LDPC. Herein, we take semantic segmentation as an example visual analytic task, employ HyperSeg \cite{hyperseg} as a powerful analytics model to generate segmentation prediction, and evaluate the capacity of our DVST on the popular CamVid benchmark \cite{CamVid1, CamVid2}. CamVid is a road scene understanding dataset, which offers four video clips of driving scenes, and part of frames are densely semantic annotated. Following the training protocol of \cite{hyperseg}, we use 468 annotated images as the training set and the other 233 ones as the test set. Each video and the labeled images are resized to $256\times 192$. To achieve a better rate-distortion-accuracy trade-off, we finetune the DVST (PSNR) model over the CamVid training set. The whole training loss extends \eqref{eq_GOP_training_loss} to achieve joint optimization of both video transmission and analysis. It is formulated as $L = \frac{1}{N}\sum_{t=1}^{N} \left( \lambda k_t + D_t + \beta D_{t,{\rm seg}} \right)$, where $D_{t,{\rm seg}}$ denotes the boot-strapped cross entropy loss \cite{reed2015training} specialized for semantic segmentation. $\beta$ is the weighting parameter, we set $\beta = 64$ to balance the importance among reconstruction and segmentation.
During the evaluation on the test set, all frames are transmitted to the receiver to calculate signal level distortion in terms of PSNR, while only the labeled frame participates in the calculation of segmentation performance (the labeled frame evenly distributes in the second or last frame of GOP). We report the class mean intersection over union (mIoU) results, a standard evaluation metric for semantic image segmentation \cite{hyperseg}. A higher mIoU score indicates a better match between prediction and ground truth, with a maximum value of 1.

The quantitative results are shown in Fig. \ref{Fig15}. We observe that our DVST achieves better performance for both reconstruction and segmentation with various CBRs. It outperforms the two separated coding schemes and the gap increases with CBR. Hence, our DVST method can better support machine vision tasks and hold higher fidelity for human vision at the same time. In addition, we present a group of segmentation example in Fig. \ref{Fig16}. The reconstruction of DVST preserves more semantic information for machine recognition, which leads to more accurate segmentation results.

\section{Conclusion}\label{section_conclusion}

This paper has proposed a new class of high-efficiency deep JSCC methods to achieve end-to-end video transmission over wireless channels. It was collected under the name ``DVST''. This DVST framework has exploited nonlinear transform and conditional coding architecture to adaptively extract semantic features across video frames, and transmit semantic features via a group of learned variable-length deep JSCC codecs and wireless channel. Benefiting from the strong temporal prior provided by the semantic feature domain context and the deep JSCC codeword domain context, the DVST framework works highly efficient and effective. The whole video transmission system design has been formulated as an optimization problem whose goal is to minimize the end-to-end transmission rate-distortion performance under established perceptual quality metrics or downstream task metrics, which well matches with the goal of end-to-end semantic communications. Extensive numerical results have shown that the proposed DVST method can generally surpass traditional wireless video coded transmission schemes. In a nutshell, this paper has proposed a promising method to attain a customized design of learning-based source-channel coding for video transmission in future semantic communications.

\bibliographystyle{IEEEbib}
\bibliography{myRef}

\end{document}